\documentclass[12pt,libertinus]{manurxiv}

\usepackage{amsmath,amssymb}

\usepackage[style=numeric,backend=bibtex]{biblatex}
\title{
Scalable semi-supervised dimensionality reduction with GPU-accelerated EmbedSOM
}
\version{Author draft}

\author[1]{Adam Šmelko}
\author[1]{Soňa Molnárová}
\author[2,$\ast$]{Miroslav Kratochvíl}
\author[3]{Abhishek Koladiya}
\author[3]{Jan Musil}
\author[1]{Martin Kruliš}
\author[4]{Jiří Vondrášek}
\affil[1]{Faculty of Mathematics and Physics of Charles University, Prague, Czech Republic}
\affil[2]{Luxembourg Centre for Systems Biomedicine, University of Luxembourg}
\affil[3]{Institute of Haematology and Blood Transfusion, Prague, Czech Republic}
\affil[4]{Institute of Organic Chemistry and Biochemistry, Prague, Czech Republic}
\affil[$\ast$]{Contact: see {\tt github.com/molnsona/blossom}}

\DeclareMathOperator*{\argmin}{\arg\min}
\def\Alg#1{{\bfseries\scshape #1}}
\def\alg#1{{\scshape #1}}
\usepackage{tikz}
\usetikzlibrary{positioning}

\def\IEEEPARstart{}

\abstract{
Dimensionality reduction methods have found vast application as visualization tools in diverse areas of science.
Although many different methods exist, their performance is often insufficient for providing quick insight into many contemporary datasets, and the unsupervised mode of use prevents the users from utilizing the methods for dataset exploration and fine-tuning the details for improved visualization quality.
We present BlosSOM, a high-performance semi-supervised dimensionality reduction software for interactive user-steerable visualization of high-dimensional datasets with millions of individual data points.
BlosSOM builds on a GPU-accelerated implementation of the EmbedSOM algorithm, complemented by several landmark-based algorithms for interfacing the unsupervised model learning algorithms with the user supervision.
We show the application of BlosSOM on realistic datasets, where it helps to produce high-quality visualizations that incorporate user-specified layout and focus on certain features.
We believe the semi-supervised dimensionality reduction will improve the data visualization possibilities for science areas such as single-cell cytometry, and provide a fast and efficient base methodology for new directions in dataset exploration and annotation.

}

\begin{document}
\maketitle

\IEEEPARstart{D}{imensionality} reduction algorithms emerged as indispensable utilities that enable various forms of intuitive data visualization, providing insight that in turn simplifies rigorous data analysis.
Various algorithms have been proposed for graphs and high-dimensional point-cloud data, and many different types of datasets that can be represented with a graph structure or embedded into vector spaces.
The development has benefited especially the life sciences, where algorithms like t-SNE~\cite{maaten2008visualizing} reshaped the accepted ways of interpreting many kinds of measurements, such as genes, single-cell phenotypes and development pathways, and behavioral patterns~\cite{toghi2019quantitative,cande2018optogenetic}.

Performance of the non-linear dimensionality reduction algorithms becomes a concern if the analysis pipeline is required to scale or when the results are required in a limited amount of time such as in clinical settings.
The most popular methods, typically based on neighborhood embedding computed by stochastic descent, force-based layouting or neural autoencoders, reach applicability limits when the dataset size is too large.
To tackle the limitations, we have previously developed EmbedSOM~\cite{kratochvil2019generalized}, a dimensionality reduction and visualization algorithm based on self-organizing maps (SOMs)~\cite{kohonen1990self}.
EmbedSOM provided an order-of-magnitude speedup on datasets typical for the single-cell cytometry data visualization while retaining competitive quality of the results.
The concept has proven useful for interactive and high-performance workflows in cytometry~\cite{kratochvil2020shinysom,kratochvil2020gigasom}, and easily applies to many other types of datasets.
Despite of that, the parallelization potential of the extremely data-regular design of EmbedSOM algorithm has remained mostly untapped.

Our contribution in this paper is a natural continuation of the development:
We describe an efficient, highly parallel GPU implementation of EmbedSOM designed to provide interactive results on large datasets.
The implementation is sufficiently fast to provide real-time visualizations of datasets larger than $10^5$ of individual data points on off-the-shelf hardware, while maintaining smooth video-like frame rate.
We demonstrate that the result gives unprecedented, controllable view of the details of specific high-dimensional datasets.
The instant feedback available to the user opens possibilities for partial supervision of the visualization process, allowing user-intuitive resolution of possible visualization ambiguities as well as natural exploration of new datasets.
We demonstrate some of the achievable results on two realistic datasets.
The resulting software, called \emph{BlosSOM}, is published as free and open source.
BlosSOM can be readily utilized to reproduce our results and explore more datasets; additionally it contains support for working with data formats (mainly, the FCS standard~\cite{fcs}) that make it immediately useful for visualization of existing and new biological data.

In the paper, we briefly describe the EmbedSOM algorithm (Section~\ref{ssec:embedsom}), and show an extension of its generalized form that dynamically mixes the user feedback to the learning process, thus enabling the semi-supervised learning (Section~\ref{ssec:dynamic}).
We specifically detail the CUDA-based GPU implementation of the algorithm in Section~\ref{sec:impl}, and report the achieved performance improvements (Section~\ref{ssec:perf}).
Finally, we showcase the achievable results on biological data, and discuss possible future enhancements and applications that would aid data analysis (Sections~\ref{ssec:appl}, \ref{ssec:future}).

\section{Background \& Related work}

In this work, we specifically reflect the needs of many areas of life sciences where large multidimensional point-cloud-like datasets occur, such as population biology, microscopy imaging, metagenomics, and others.

Single-cell cytometry~\cite{adan2017flow} forms a canonical example of this niche:
The recent development of hardware and measuring equipment has enabled precise collection of multiple features of each of millions of cells in a sample.
Clinicians and biologists commonly measure metrics such as protein expression on cell surface using antibody-based markers.
For instance, the marker detection may be performed optically by exciting fluorochromes with a laser and measuring emission spectra or using specific binding of heavy-metal ions that are detected by mass spectrometry techniques such as time-of-flight~\cite{spitzer2016mass}.
Both methods allow cheap acquisition of data about more than 50 selected features at once, typically with several millions single-cell measurement from each sample~\cite{vanikova2021omip,rodriguez2020systems}.
Additionally, the development of single-cell sequencing methods has enabled to sequence mRNA present in the individual cells~\cite{ziegenhain2017comparative}, typically yielding a dataset with thousands of data points of dimension higher than $10^4$.

Due to the variability in the samples, data, and measurements, the analysis and interpretation of the results is challenging.
Biologists usually choose to analyze the datasets by linear projection of the features to 2-dimensional plots, and selecting the cells of interest manually in a process called gating~\cite{bashashati2009survey}.
While computationally simple and easily interpretable by humans, gating gets extremely error-prone as the dimensionality of the dataset increases and it does not provide good support for detection of dataset features of dimensionality higher than $2$, such as complicated pathways and loops in cell phenotypes.
Similarly, application of algorithms for clustering analysis provided good detection of cell phenotype clusters, but minimal reliability in pathway-style feature detection~\cite{saelens2018comparison}.

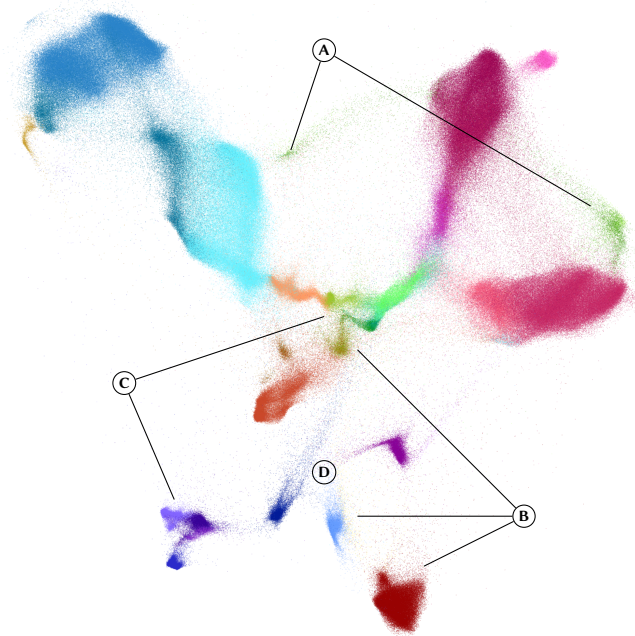
\begin{figure}
\centering
{\linewidth=21pc
\begin{tikzpicture}[font=\tiny\sffamily\bfseries, inner sep=1pt]
\node[inner sep=0, anchor=south west] (img) at (0,0) {\includegraphics[width=\linewidth]{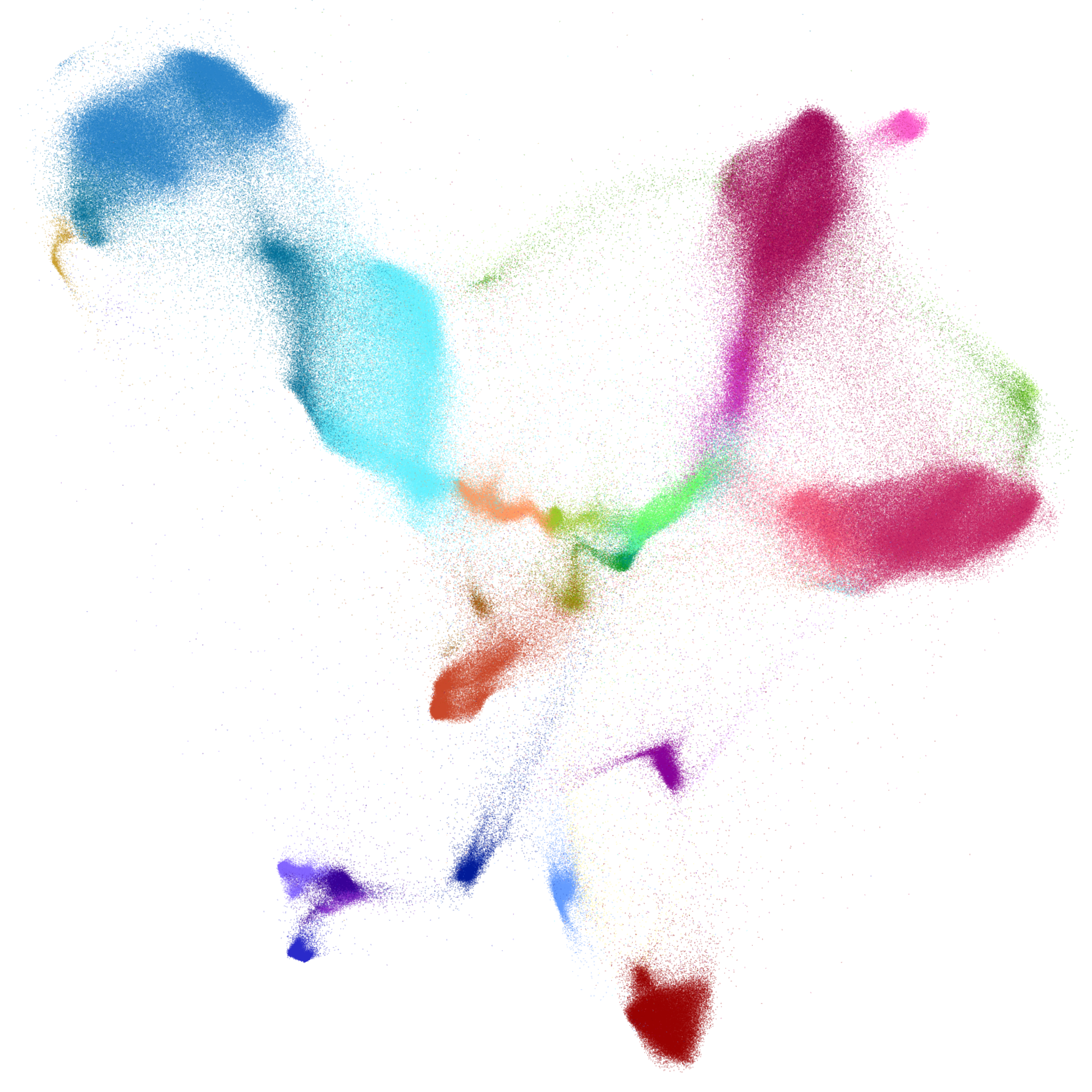}};
\node[circle, draw] (A) at (0.5\linewidth,0.9\linewidth) {A};
\node[circle, draw] (B) at (0.8\linewidth,0.2\linewidth) {B};
\node[circle, draw] (C) at (0.2\linewidth,0.4\linewidth) {C};
\node[circle, draw] (D) at (0.5\linewidth,0.266\linewidth) {D};
\draw (A) to (0.45\linewidth,0.75\linewidth);
\draw (A) to (0.9\linewidth,0.666\linewidth);
\draw (B) to (0.55\linewidth,0.2\linewidth);
\draw (B) to (0.65\linewidth,0.125\linewidth);
\draw (B) to (0.55\linewidth,0.45\linewidth);
\draw (C) to (0.5\linewidth,0.5\linewidth);
\draw (C) to (0.275\linewidth,0.225\linewidth);
\end{tikzpicture}}
\caption{
Example EmbedSOM projection of 841,644 data points with 39-dimensional single-cell measurements, representing the immune cell contents of bone marrow~\cite{samusik2016automated}.
Colors were assigned manually to differentiate biologically relevant cell populations.
Manual intervention in the unsupervised dimensionality reduction process would allow the user to fix several visualization deficiencies:
overlapping pathways (labeled with~A), disconnected pathways (B), display of features in the small complex cell clusters (C), and the orientation and positioning of clusters that was chosen arbitrarily by the reduction process, not reflecting any biologically relevant features (D).
}
\label{fig:samusik}
\end{figure}

The development of non-linear dimensionality reduction algorithms in the last two decades has resolved a large set of these challenges. The new visualization-oriented methods, represented in the field mainly by t-SNE~\cite{maaten2008visualizing}, provided output that was sufficiently intuitive to understand, yet gave a satisfactory view of the highly complicated features that can not be observed by gating and projections.
The tremendous success was quickly followed by new alternative algorithms that optimize different aspects of the process.
UMAP~\cite{becht2019dimensionality} is currently a common choice for both visualization and a starting point of analysis, followed by PHATE~\cite{moon2019visualizing}, scvis\cite{ding2018interpretable}, TriMap\cite{amid2019trimap}, and others.
For illustration, a typical example visualization is provided in~Figure~\ref{fig:samusik}, along with the description of common visualization problems.

Following the plethora of newly introduced algorithms, a discussion unfolded to assess optimality of the obtained visualization.
Reviews have focused not only on reproducibility and robustness (i.e., susceptibility to large changes in output caused by small variations in data or different random seed), but also on representation of biologically valid features.
Vast resources were invested into modifying the algorithms, especially t-SNE, to maximize various metrics, including convergence speed~\cite{belkina2019automated}, general performance~\cite{linderman2019fast}, robustness~\cite{polivcar2021embedding}, and the (very informally specified) quality of display of local and global relations in data~\cite{kobak2019heavy,kobak2021initialization}.
Some of the features required for high visualization quality are showcased in~Figure~\ref{fig:samusik}.
Still, despite of many GPU acceleration efforts, including t-SNE~\cite{chan2018t} and UMAP~\cite{nolet2020bringing}, most of the algorithms suffer from performance limitations that effectively prevent interactive real-time applications.

User interaction possibilities in the dimensionality reduction process were therefore largely neglected, except for prohibitively small datasets where the use of force-based graph layouts and similar algorithms did not pose a throughput challenge.
As one of few exceptions, van~Unen~et~al.~\cite{unen2017visual} and Chatzimparmpas~et~al.~\cite{chatzimparmpas2020t} achieved a methodological advance with extending the t-SNE algorithm, producing respectively HSNE and t-viSNE.
HSNE organizes and visualizes a small model of data dynamically using t-SNE, and provides an intuitive way for the user to zoom into various compartments of the dataset, following a hierarchical structure of clusters.
t-viSNE focuses on interactive use of t-SNE for exploration of complex dataset properties, but only of relatively small datasets.

Compared to HSNE, our developments in BlosSOM provide two major improvements:
Full dataset may be rendered at all times (giving an unprecedented high-definition view of the features), which is enabled by more efficient design of the base algorithm.
Additionally, no hierarchical structure or no fixed layouting algorithm is imposed to the user, improving the display of structures that are hard to visualize or capture with hierarchical methods, such as the inter-cluster pathways.

\section{Methods\label{sec:methods}}

\subsection{Landmark-directed dimensionality reduction}
\label{ssec:embedsom}

EmbedSOM is a visualization-oriented method of non-linear dimensionality reduction that works by describing a high-dimensional point by its location relative to landmarks equipped with a topology, and reproducing the point in a low-dimensional space using an explicit low-dimensional projection of the landmarks with the same topology~\cite{kratochvil2019generalized}.
The ability to effectively work with a simplified model of the data differentiates it from other dimensionality reduction methods; in turn it offers superior performance by reducing the amount of necessary computation as well as by opening parallelization potential, since the computations of the projections of many individual points are independent.
In the setting of flow and mass cytometry data visualization, this provided speedup of several orders of magnitude against the other available methods~\cite{kratochvil2020gigasom,kratochvil2020shinysom}.

While the EmbedSOM originally used the (eponymous) self-organizing maps to find the viable high- and low-dimensional manifolds from the data points, the concept generalized well to many other methods.
In particular, the projection has been shown to work with any (even random) set of high-dimensional landmarks that have the low-dimensional counterparts organized by any selected dimensionality reduction method (which may be slow in comparison, given the fact that the set of landmarks is usually small).
In Section~\ref{ssec:dynamic}, we utilize this freedom of model specification to provide dynamic view of the dataset, based on a simplified dataset model that the user may refines in order to improve the dataset view.

More formally, the EmbedSOM algorithm works as follows: We take $d$ to be the dimension of the high-dimensional space and assume the low-dimensional space to be 2-dimensional for brevity. EmbedSOM processes $n$ $d$-dimensional points in a matrix $X$ of size $n\times d$, and output $n$ 2-dimensional points in matrix $x$ of size $n\times 2$.
The high- and low-dimensional landmarks similarly form matrices $L$ of size $g\times d$ and $l$ of size $g\times 2$, where usually $g\ll n$.
Each point $X_i$ is transformed to a point $x_i$ as follows:
\begin{enumerate}
\item $k$ nearest landmarks are found for point $X_i$ ($k$ is a constant parameter satisfying $3\leq k\leq g$)
\item the landmarks are ordered and a score is assigned to each of them, using a smooth function of the distance that assigns highest score to the closest landmark and $0$ to the $k$-th landmark (this ensures the smoothness of projection in cases when $k<g$ \cite[Methods section]{kratochvil2019generalized})
\item for each pair $(u,v)$ of the closest $k-1$ landmarks (i.e., the ones with non-zero score), a projection of the point $X_i$ is found on the 1-dimensional affine space with coordinate 0 at $L_u$ and 1 at $L_v$; the 1-dimensional coordinate of the projection in this affine space is taken as $D_{uv}(X_i)$ and the same projected coordinates are defined in the low-dimensional space as $d_{uv}(x_i)$
\item point $x_i$ is fitted to the low-dimensional space so that the squared error in the coordinates weighed by nearest-landmark scores is minimized: $$x_i = \argmin_{p\in \mathbb{R}^2} \sum_{u,v}s_u\cdot s_v\cdot \left(D_{uv}(X_i)-d_{uv}(p)\right)^2$$
\end{enumerate}
Because $d_{uv}(p)$ is designed as a linear operator, the error minimization problem in the last step collapses to a trivial solution of $2$ linear equations with $2$ variables.
Full algorithm pseudocode may be found in the original publication~\cite[Algorithm 1]{kratochvil2019generalized}.

Efficient implementation of the EmbedSOM algorithm is the main performance concern that enables its interactive use.
The original CPU-based parallel implementation was able to visualize hundreds of thousands of points per second on common use-cases.
As a major result of this paper, in Section~\ref{sec:impl} we improve this performance to the scale of milliseconds, enabling real-time projection and rendering of points based on interactive control of the high- and low-dimensional landmarks.

\subsection{User supervision and model interaction}
\label{ssec:dynamic}

EmbedSOM landmarks (the matrices $L$ and $l$) represent a simplified dataset model that can be used to conveniently and predictably steer the dimensionality reduction.
In particular, the main property of the projection --- visualizing the data points from the neighborhood of a landmark $L_i$ preferably in the neighborhood of the corresponding low-dimensional $l_i$ --- gives an intuitive interpretation for the landmark positions:
Manipulating the high-dimensional landmarks chooses which data are visualized, while manipulating the low-dimensional landmarks chooses the desired location of the visualized points.
Smoothness of the projection then grants that the smooth manipulations of the landmarks that will result in smooth changes of the results, enabling predictable user control and refinement.

However, positioning of the landmarks in the high-dimensional space (which is inherently complicated to navigate) and finding a suitable layout of the landmarks in the low-dimensional space is an overly complicated task for the user alone.
The main concern of this section is to design a simplification of the control of the landmarks, so that viable results may be reached in an automated way and the user interaction is required only for decisions that can not be decided automatically such as resolving dimensionality-reduction ambiguities and positioning of the dataset parts that matches some assumed semantics.
We describe two ways of automated and user-controlled positioning of the landmarks that implement this kind of partial supervision, thus making the method semi-supervised.
Both are roughly based on the embedding methods proposed in previous work~\cite{kratochvil2019generalized}; only modified for interactive environment.

The main tasks that the user supervision interface has to resolve are thus as follows:
\begin{itemize}
\item place the landmarks to viable positions in the high-dimensional space
\item dynamically increase or decrease the resolution of the model in specified places, by adding or removing landmarks
\item organize low-dimensional landmarks to reflect the structure in the high-dimensional space, while allowing the user to resolve ambiguities that arise in dimensionality reduction
\item react to the changes in the input datasets, such as scaling of the dimensions and appearance of new points
\end{itemize}

\begin{figure}
\centering\includegraphics[width=\linewidth]{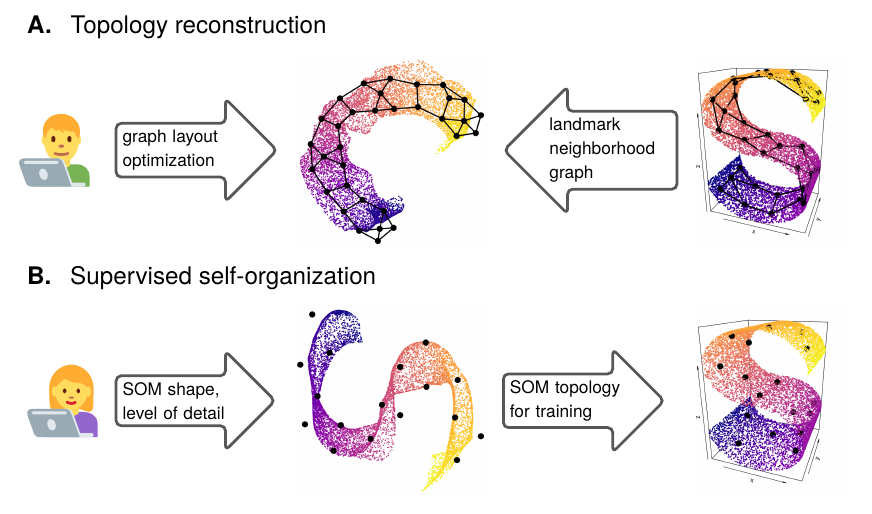}
\caption{
Schema of the 2 implemented user supervision approaches.
The user interacts with low-dimensional model that is intuitive to navigate, and indirectly drives the positioning of the corresponding high-dimensional image of the model, using either a graph-based approach (Section~\ref{sssec:sup-graph}) or self-organizing map approach (Section~\ref{sssec:sup-som}).
}
\label{fig:supervis}
\end{figure}

The two methods detailed below are briefly illustrated in Figure~\ref{fig:supervis}.

\subsubsection{Semi-supervised structure reconstruction}
\label{sssec:sup-graph}

One possible approach to position the high-dimensional landmarks is to sample them randomly from the distribution of the original data set, and position the low-dimensional ones to reflect the structure of the sampling.
In the original unsupervised implementation, we used a random sample of the input dataset as the landmark positions, and a dimensionality reduction methods such as t-SNE to position the landmarks.
Importantly, the positioning of low-dimensional landmarks could be performed relatively quickly even by rather time-demanding algorithms such as t-SNE, because the algorithm only had to work with the highly reduced version of the dataset in landmarks.

In the dynamic, supervised context, we need to avoid the randomness in order to avoid flickering in the view of the dataset, and utilize a dimensionality reduction algorithm that may reflect the user input.
Thus, we chose to continuously run an interactive version of $k$-means clustering with a low learning rate to find good $k$ high-dimensional landmarks $L$, and employ a simple force-based graph layouting algorithm on a neighborhood graph of $L$ to embed the landmarks to 2D.
Both these algorithms are capable of smooth transitions between consecutive states, thus avoiding the flicker.
Moreover, force-based graph layouting may be intuitively steered by the user by dragging the graph nodes. Similarly, the points may be added and removed from $k$-means clustering in the same interface in order to increase and decrease the model resolution.

Addition of a new landmark is implemented as follows: The user selects a low-dimensional landmark, and upon pressing a special button, both the low-dimensional landmark and its high-dimensional counterparts are duplicated in their respective spaces.
The $k$-means algorithm then consecutively optimizes the positions of the landmarks to provide a detailed view.
This stability of the result is helped by the initialization properties of $k$-means where the cluster centroids tend to stay in the same clusters~\cite{franti2019much} (counter-intuitively, the same properties have an undesirable impact on the robustness of unsupervised clustering).
Most importantly, this enables the user to position new landmarks without having to navigate the possibly overwhelming complexity of the data distribution in the high-dimensional space.

\subsubsection{Supervised training of self-organizing maps}
\label{sssec:sup-som}

Alternatively, the user may choose a SOM approach as originally intended for EmbedSOM.
BlosSOM supports user drawing of the 2-dimensional version of the SOM on a canvas, which is used as-is as the low-dimensional landmarks.
New landmarks may be added at any position, as their initial high-dimensional coordinates can be fitted using the coordinates of the other close landmarks in 2D.

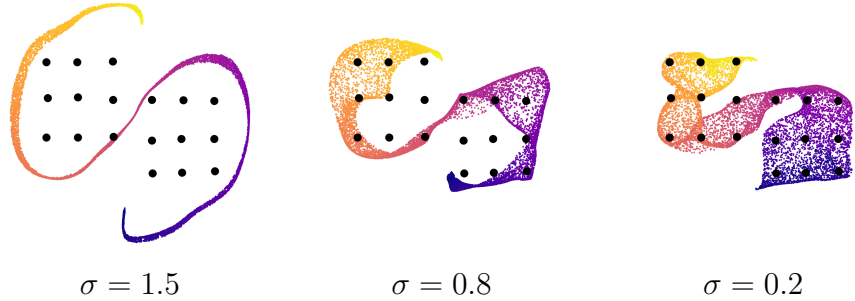
\begin{figure}
\centering
\begin{tikzpicture}[node distance=1em]
\node (a) {\includegraphics[width=.25\linewidth]{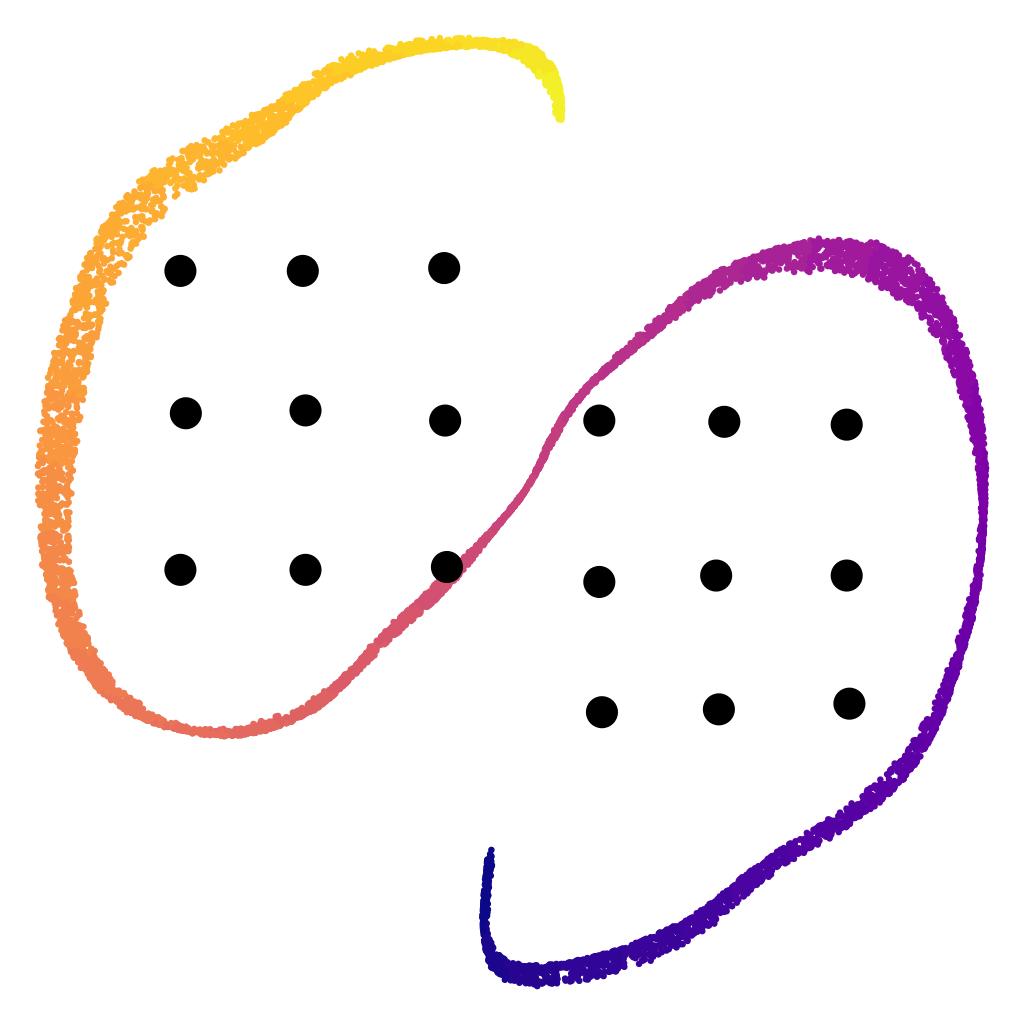}};
\node[right=of a] (b) {\includegraphics[width=.25\linewidth]{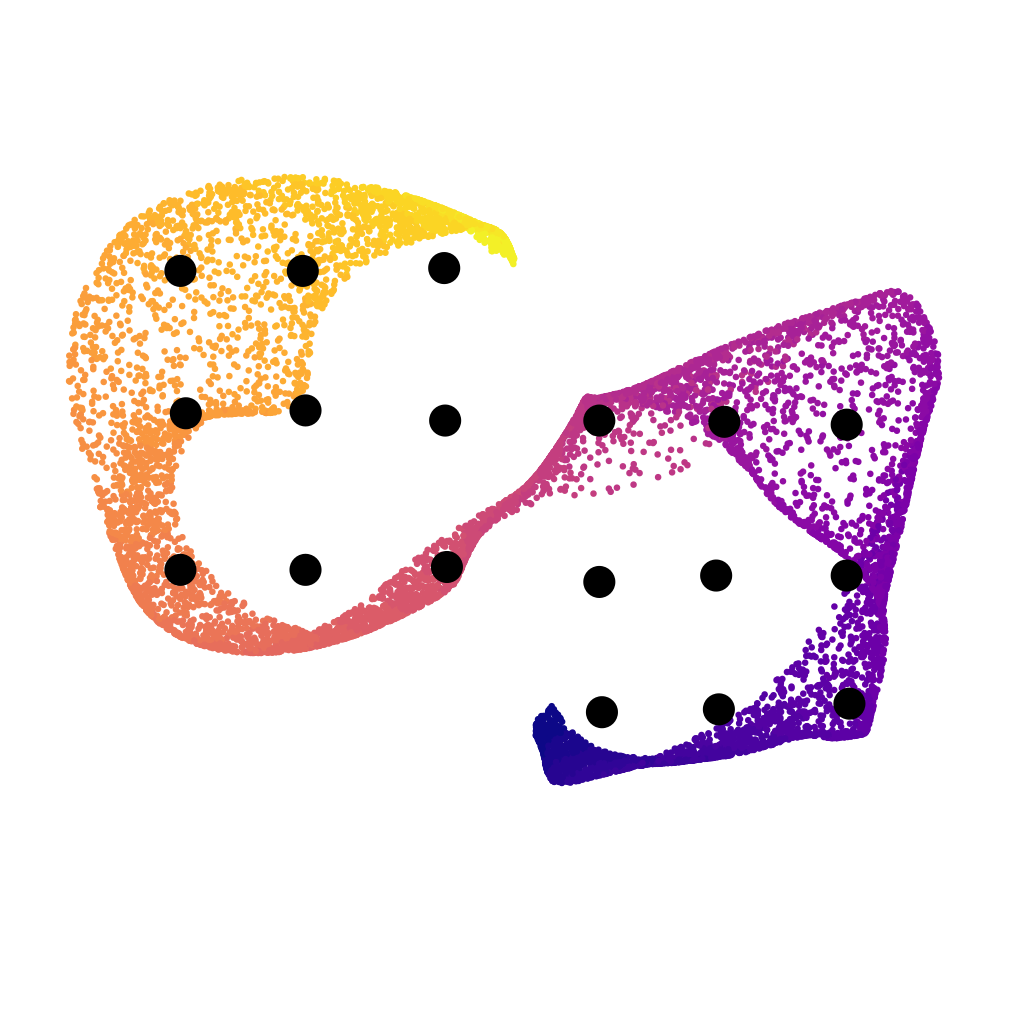}};
\node[right=of b] (c) {\includegraphics[width=.25\linewidth]{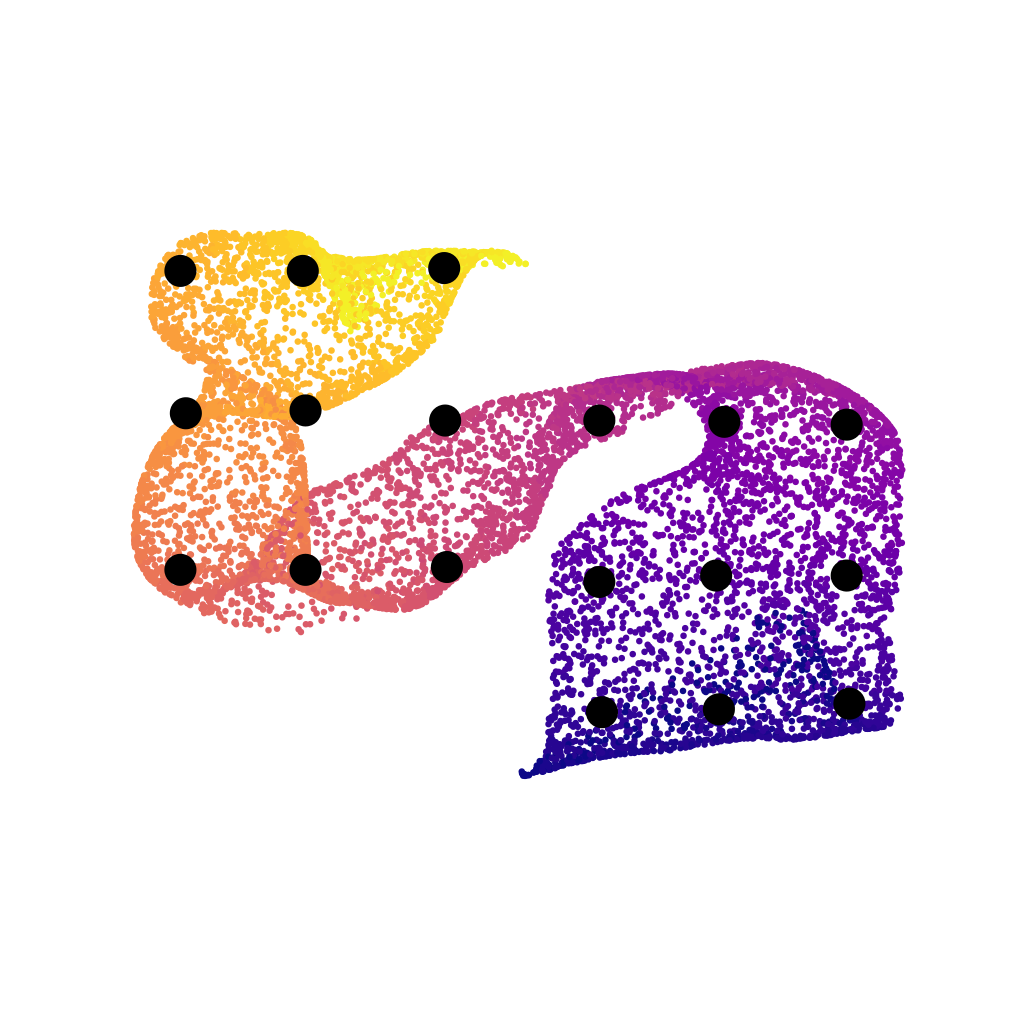}};
\node[below=0pt of a] {$\sigma=1.5$};
\node[below=0pt of b] {$\sigma=0.8$};
\node[below=0pt of c] {$\sigma=0.2$};
\end{tikzpicture}
\caption{
Effect of different settings of $\sigma$ SOM training parameter on the output detail, demonstrated on an extruded S-shaped 3D dataset and a custom SOM shape.
Value of $\sigma$ decreases from left to right, progressively revealing finer details but losing larger-scale structure.
}
\label{fig:somsigma}
\end{figure}

The positioning of the landmarks in 2D is then used as a topology for training the high-dimensional landmarks as neuron weights of the SOM algorithm.
To extend the supervision possibilities of this step, BlosSOM adds specific controls that allow the user to manually sweep through the SOM neighborhood sizes and learning rates (usually labeled $\sigma$ and $\alpha$ \cite{kohonen1990self}), which is done automatically in unsupervised SOM training.
This allows the users to optionally pause the SOM training at any stage and fix or customize the SOM topology at coarse detail level (with larger $\sigma$) before it is used to train fine details (small $\sigma$, the difference is closer detailed in Figure~\ref{fig:somsigma}).

\section{GPU implementation of EmbedSOM}
\label{sec:impl}

While EmbedSOM is relatively straightforward to parallelize for mainstream CPU architectures, several challenges appear in the design of an optimal implementation for contemporary GPUs.
In this section we outline the key optimizations that allowed us to run the high-performance dimensionality reduction in BlosSOM, and give an overview of the relative performance gains achieved by the algorithm choice.

Technically, the algorithm consists of two main parts that provide distinct implementation challenges:

\begin{itemize}
\item {\bfseries $k$-NN step}\quad
The selection of $k$ nearest neighbors in $L$ for each point from the dataset $X$ requires a highly irregular selection of indices of $k$ lowest values from columns of the dynamically computed distance matrix $L^T\cdot X$.
In Section~\ref{sec:impl-knn}, we describe a way to utilize partial bitonic sorts to run this step in parallel with minimal divergence.

\item {\bfseries Projection step}\quad
Computation of the small linear system that is used to find the minimal-error-projection of a point, namely of projections $D_{uv}$ and the derivatives $\frac{\delta d_{uv}}{\delta x_i}$ (see Section~\ref{ssec:embedsom}), is hard to parallelize due to irregular memory access patterns while collecting the data for the computation.
In Section~\ref{sec:impl-projection}, we describe an arrangement of computation that utilizes shared memory, thread grouping, and privatized accumulators for minimizing the memory latency and improving the throughput of the operation.
\end{itemize}

We carried out the implementation in NVIDIA CUDA~\cite{guide2013cuda}, the performance validation presented in the next sections is accordingly carried out only on NVIDIA hardware that supports CUDA.
We used 3 different generations of NVIDIA GPUs to gather a comprehensive picture of the influence of the optimizations on hardware of different power.
Details about the benchmarking hardware and data collection are available in Section~\ref{ssec:bench}.

Despite our benchmarks are NVIDIA-specific, the presented kernels do not depend on any NVIDIA-specific functionality, and the results should be portable to other GPU programming frameworks (such as Vulkan Compute shaders) and to hardware of other vendors.
We expect that only minor adjustments will be required to compensate for GPU design differences, such as the 64-thread wavefronts on AMD devices.

\subsection{Benchmarking setup and methodology}\label{ssec:bench}

The main objective of the benchmarking was to measure the speedups achieved by different applied optimizations, in order to determine the optimal algorithms and parameter setting for the sub-tasks of EmbedSOM computation.

The timing results, presented in the following sections, were collected as kernel execution times measured by standard system high-precision clock available on the platforms.
We tested that the relative standard deviations in the 10 collected measurements of each result were less than 5\% of the mean value, and report the mean values.
For brevity, in this paper we report only the results from Tesla~V100~SXM2; all benchmark results are available in the online repository (Section~\ref{ssec:data}).

The precise platforms used for collecting the results are as follows:
\begin{itemize}
	\item NVIDIA GeForce GTX~980 (Maxwell architecture, compute capability 5.2, clocked at $1.2$GHz), running Rocky Linux 8
	\item NVIDIA GeForce RTX 3070 (laptop version, Ampere architecture, compute capability 8.6, clocked at $1.6$GHz), running Windows 10 (build 19043.1237)
	\item NVIDIA Tesla V100~SXM2 (Volta architecture, compute capability 7.0, clocked at $1.3$GHz), running Rocky Linux 8.
\end{itemize}
Benchmarks of the serial performance of the CPU implementation were collected on Intel Xeon Gold 5218 clocked at 2.3GHz (on the same machine as the Tesla~V100~SXM2 benchmarks).
We used CUDA toolkit version 11.4 on all machines.

Notably, the tested configurations span 5 generations of NVIDIA hardware (skipping Pascal (CC 6) and Turing (CC 7.5) architectures), and all three commonly utilized types of hardware (laptop, hi-end desktop, server).
Since all architectures exhibited similar relative speedups between algorithms and configurations, we expect that our assumptions in the optimization process are equally valid for most other modern GPUs, except perhaps for specialized low-power laptop and mobile hardware.

All benchmarking datasets were synthetic, all containing exactly $1$Mi points ($n=2^{20}$, reflecting the common sizing of real-world datasets~\cite{adan2017flow}) with all coordinates sampled randomly from the same uniform distribution.
Performance of the benchmarked algorithms is not data-dependent, except for the case of caching performance in the projection step, where the completely random dataset is a worst-case scenario.

\subsection{$k$-NN selection step}\label{sec:impl-knn}

The task of the first part of the algorithm is to find $k$ nearest landmarks (from $L$) for every data point in $X$.
This comprises two sub-steps: computing Euclidean distances for every pair from $L$ and $X$, and performing point-wise reduction that selects a set of $k$ nearest landmarks for each of the $n$ points, based on the computed distances.

While the Euclidean distance computation is mathematically simple and embarrassingly parallel, achieving optimal throughput on GPUs is quite challenging~\cite{krulivs2017employing}.
In particular, the ratio between the data transfers and the arithmetic operations performed by each GPU core is heavily biased towards data transfers (with respect to the contemporary GPU parameters).
The overhead of data transfers is best prevented by finding a good caching pattern for the input data that is able to optimally utilize all hardware caches (L1 and L2), shared memory, and core registers.

The parallel implementation of $k$-nearest neighbors search is even more challenging.
The $k$-NN problem is computed individually for each data point, which provides the space for possible parallelization.
However, concurrently processed instances of a na\"{i}ve $k$-NN implementation exhibit severe code divergence because the selection process is purely data-driven, and require a high amount of memory allocated per core.
Optimally, the $k$-NN selection is realized by customized versions of parallel sorting algorithms, which are well researched and possess existing GPU implementations~\cite{singh2018survey}.

Our implementation chooses to optimize both sub-steps, since the ratio of the amount of required computations can be easily biased by configuration of parameters $d$ and $k$.
In particular, processing high-dimensional datasets with a low $k$ parameter spends significantly more time in the distance computation, but lower-dimensional datasets with higher $k$ require more time in the nearest neighbor selection.

As another concern, the implementation may use separate kernels for both sub-tasks, or a single fused kernel.
Kernel separation provides better code modularity and much flexibility in work-to-thread division and data caching strategy, at the cost of having to materialize all the computed distances in the GPU global memory, thus significantly increasing the total amount of data transfers.
Contrary to that, a fused kernel may immediately utilize the computed distances in $k$-NN computation without transferring the data to global memory, and interleaving of the distance computations with $k$-NN may help to improve the ratio between computations and data transfers.
Since our initial observations showed that the overhead of the data transfers required for kernel communication is relatively high, we decided to implement only the fused variant for the sake of simplicity.
Usage of separate kernels might be interesting in the future, especially for extreme values of $d$ that diminishes the relative cost of the distance data transfer.

\subsubsection{Available algorithms for $k$-NN}

There are many approaches to $k$-NN selection, varying in complexity and parameter-dependent performance.
To substantiate our choice of the algorithm for GPU EmbedSOM, we have implemented and benchmarked several of the possibilities, as described further in this section.

As a baseline (labeled \Alg{Base}), we used the most straightforward approach to GPU parallelization, implemented in a similar manner as the sequential CPU implementation.
The \alg{Base} kernel is spawned in $n$ threads (one for each data point), and each thread computes the distance between its data point and all landmarks, while maintaining an ordered array of $k$ nearest neighbors.
The array is updated by an insert-sort step performed for every new computed distance --- i.e., by starting at the end of the array and moving the new distance-index pair towards smaller values until it reaches the correct position.

\Alg{Shared} algorithm is a modified version of the baseline algorithm that utilizes shared memory as a cache, following the recommended optimization practice of improving performance by caching data that are reused multiple times~\cite{guide2013cuda}.
In this case, we cache the landmark coordinates, which are sufficiently small to fit in the shared memory for all tested parametrizations.

In \Alg{GridInsert} algorithm, we utilize the shared memory to cache both landmarks and points.
However, the size of shared memory is very limited, forcing us to choose a good amount of the data to cache; we thus parametrized the algorithm by block height $h$ (number of cached points from $X$) and block width $w$ (number of cached landmarks from $L$).
With this, the algorithm runs in epochs, each of which first caches the points and landmarks, and computes $h \cdot w$ distance values using only the cached data.
While the distances are computed concurrently by the whole thread block, we chose to avoid explicit synchronization in the $k$-NN step, using only $h$ threads to incorporate the newly computed distances into $h$ separate $k$-NN results using the insert-sort steps.
We therefore expected \alg{GridInsert} to achieve better throughput in the distance computation thanks to the caching, at the cost of slightly sub-optimal $k$-NN reduction, thus giving best performance on high-dimensional datasets and low values of $k$.

The above algorithms focus solely on optimizing the distance computation; we further detail the possible optimizations of the $k$-NN selection.

A straightforward way for computing the $k$ nearest neighbors in parallel is to sort an entire array of distances using a parallel sorting algorithm, then taking the first $k$ items.
Although the overhead of storing the distances might be excessive, we expected the strategy to be competitive especially for large values of $k$ (approaching the total number of landmarks in the grid).
We use this approach in \Alg{Radix} algorithm, which employs the highly-optimized sorting algorithm from the state-of-art CUB library~\cite{cub}.
The algorithm allocates an entire thread block to process one input data point.
The block cooperates on computing the Euclidean distances by dividing the landmarks evenly among the threads.
The distances are stored along with indices in the shared memory block, which is then sorted by the CUB radix sort, and subsequently the first $k$ items are copied to the result buffer in global memory.
Importantly, the whole block of $g$ distance-index pairs must fit in the shared memory, which imposes a limitation on the maximal amount of landmarks, and prevents much of the input caching in the shared memory, impacting the efficiency of distance computation.

Finally, improvising on our previous work~\cite{krulivs2015optimizing}, we implemented \Alg{Bitonic} $k$-NN selection algorithm, which utilizes routines from the highly parallelizable bitonic sorting algorithm.
Bitonic sorting is very suitable for parallel lockstep execution~\cite{krulivs2017employing}, and the capability to merge sorted sequences has allowed us to keep only $2k$ distances (instead of $g$) in the shared memory.
Because this method benchmarked best on average and is used in BlosSOM by default, we describe it in more detail in the following section.

\subsubsection{Bitonic approach to $k$-NN}

The \alg{Bitonic} approach can be seen as a combinations of the benefits of the other algorithms:
It provides better performance than \alg{Radix} approach in the terms of selecting the $k$-NN items because it does not require to materialize all distances in the memory and do a full sort.
Even though it does not use an elaborate input caching strategy like \alg{GridInsert}, it still gives interesting results, because as the data loading operations can be partially overlapped with bitonic sorting operations if enough warps allocated to one streaming multiprocessor.

\begin{figure}
	\centering
	\includegraphics{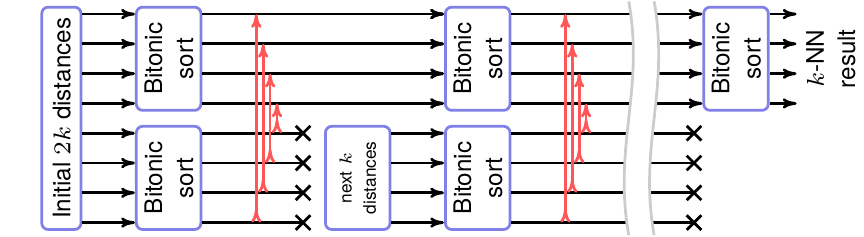}
	\caption{Schema of the $k$-NN selection in the \alg{Bitonic} algorithm (for illustration, $k=4$). Each horizontal line represents a data item in the shared memory. Red lines represent a bitonic merge that selects intermediate $k$ `best' neighbors, enabling the algorithm to discard $k$ others.}
	\label{fig:bitonic-schema}
\end{figure}

The bitonic comparator network operation provides a building block that, given two buffers of size $k$ of neighbor distances sorted by bitonic sort, selects the closest $k$ of the neighbors in a single (parallel) operation, allowing us to quickly discard neighbors that do not belong into the $k$-neighborhood.
Applying this operation iteratively on $k$-sized blocks of distances sorted by the bitonic sort (as shown in Figure~\ref{fig:bitonic-schema}), we obtain a highly performing scheme that requires only $2k$ items present in the shared memory.
In particular, the shared memory always contains a $k$-block of distances (and corresponding indexes) that holds $k$ so-far-nearest neighbors, and one block of $k$ distances that are computed from $L$; in each iteration both blocks are sorted by the bitonic sorter in parallel and merged by bitonic comparator to move the distances of new nearest $k$ neighbors into the intermediate block.
The other block is then re-filled by a new set of $k$ distances from $L$.

Technically, each step of the sorting net requires $\frac{k}{2}$ comparators, thus optimally $\frac{k}{2}$ threads that work concurrently on the $b$-sized block.
Hence, we allocate $k$ threads for each data point, which alternate their work between computing a block of $k$ distances, and performing two bitonic sorts on the 2 $k$-sized blocks in parallel.
For simplicity, our implementation assumes that $k$ is always a power of $2$, excessive output of the sorter is discarded.

\subsubsection{Parameters and performance of $k$-NN selection}\label{sec:knn-evaluation}

Here we give an overview of performance and viable parameter settings observed for the $k$-NN selection algorithms.

Notably, all algorithms for $k$-NN are affected by CUDA thread block sizing which determines warp scheduling parameters and reuse possibilities of the shared-memory cache.
We observed that total thread block size of $256$ threads was either optimal or near to optimal for almost all tested configuration, with the exception of \alg{Radix} algorithm that performed better with $128$ threads, and \alg{GridInsert} that performed the best with $64$ threads for lower values of $d$ and $g$ parameters.

Parameters $w$ and $h$ of the \alg{GridInsert} algorithm determine the ratio between data transfers and computations, but may also affect the pressure on the shared memory. (Technically, parameter $h$ is determined by the thread block size divided by $w$, we thus optimize only $w$.)
Empirical evaluation indicate that the algorithm performs best when each parallel insertion sort is performed in a separate warp, so the code divergence in SIMT execution is prevented (i.e., $w$ is a multiple of $32$)
In fact, the optimal performance was observed for $w$ equal to $3\times 32$ or $4\times 32$; however the speedup over $w=32$ is relatively low.

\begin{figure}
	\centering
	\includegraphics[width=21pc]{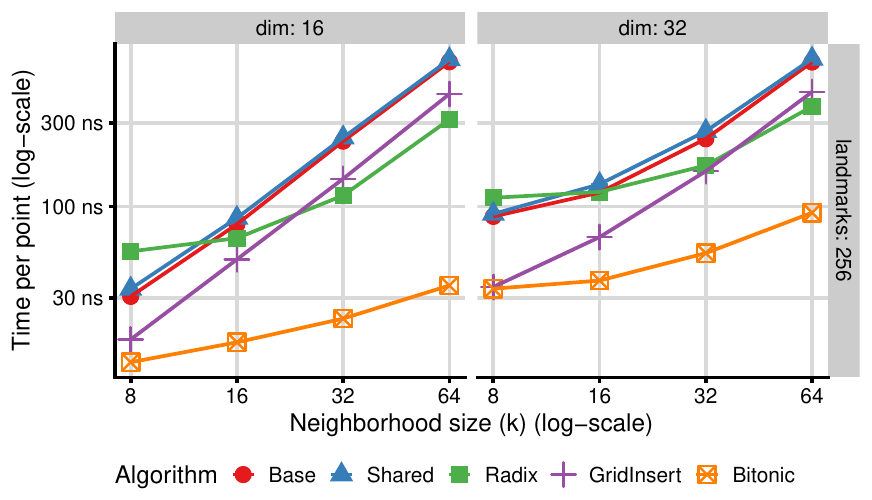}
	\caption{Amortized performance of $k$-NN step for a single input point using parameters usual in flow cytometry}
	\label{fig:knn-result-repre}
\end{figure}

A comparison of the best configurations of each algorithm on dataset sizes that are common in our target use-cases is shown in Figure~\ref{fig:knn-result-repre}.
The \alg{Bitonic} algorithm significantly outperformed the other algorithms followed by \alg{GridInsert} and \alg{Radix}, depending on the actual parametrization.
The speedup of \alg{Bitonic} over \alg{Base} was between $3\times$ to $20\times$ and usually more than $2\times$ over the second-ranking method.
Accordingly, we use only \alg{Bitonic} in the interactive version of BlosSOM.

The benchmarking also confirmed rather huge scaling difference between algorithms based on divergent insertion sort and algorithms based on sub-quadratic parallelizable sorting schemes.
We conclude that despite the simplicity that might enable GPU speedups in certain situations, the insertion sort is too slow for larger values of $k$ in this case.

As an interesting result, we observed that despite following the general recommendations, the straightforward use of shared memory (in the \alg{Shared} algorithm) did not improve overall performance over the \alg{Base}.
Quite conversely, the overhead of explicit caching even caused slight decrease in the overall performance.

\begin{figure}
	\centering
	\includegraphics[width=21pc]{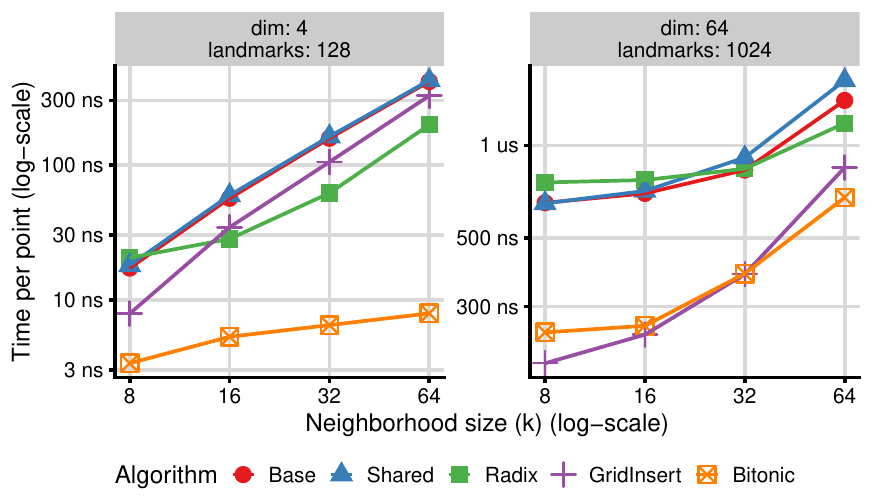}
	\caption{Amortized $k$-NN step performance in corner-case parametrizations}
	\label{fig:knn-result-extreme}
\end{figure}

We additionally report the performance measurements for two selected corner cases with extreme values of $g$ and $d$, as shown in Figure~\ref{fig:knn-result-extreme}.
Mainly, the total volume of the computation required to prepare the Euclidean distances scales with $g\cdot d$, which becomes dominant when both are maximized.
At that point, we observed that \alg{GridInsert} provides comparable or mildly better performance than \alg{Bitonic}, especially in cases where $k$ is small and the overhead of insertion sorting is not as pronounced.

Naturally, we should ask whether it could be feasible to combine the benchmarked benefits of \alg{GridInsert} and \alg{Bitonic} algorithms in order to get the best of both approaches (optimal inputs caching and fast $k$-NN filtering).
While an investigation of this possibility could be intriguing, we observed that a fused algorithm would require very complicated management of the shared memory (which both algorithms utilize heavily), and the estimated improvement of performance was not sufficient to substantiate this overhead; we thus left the question open for future research.

\subsection{Projection step}\label{sec:impl-projection}

The second part of the dimensionality reduction method is the actual projection into the low-dimensional space.
The computation of the low-dimensional point position $x_i$ by EmbedSOM involves

\begin{enumerate}
	\item conversion of the distances collected in the $k$-NN to scores,
	\item orthogonal projection of $X_i$ to $k \choose 2$ lines generated by the $k$ neighbors to create contributions to the final approximation matrix,
	\item solution of the resulting small linear system using the Cramer's rule.
\end{enumerate}

Since the first and last step are embarrassingly parallel problems with straightforward optimal implementation, we focus mainly on the orthogonal projections, where the computation is complicated by a highly irregular pattern of repeated accesses to an arbitrary $k$-size subset of $L$, and runs $\mathcal{O}(k^2)$ operations on vectors of size $d$ for each input point.

As with the $k$-NN step, we designed several algorithms that successively optimize the access patterns, detailed below.

The baseline algorithm \Alg{Base} uses the most straightforward parallel approach, where each thread computes the projection of one single point sequentially, and the concurrency is achieved only by processing multiple points simultaneously.
All data is stored in global memory, and no explicit cache control is performed.

Because the irregular repeated access to the elements of $L$ impairs the performance of the baseline algorithm, we chose to reorganize the workload.
In algorithm \Alg{Shared}, each projection is computed by a whole block of threads that cooperatively iterate over landmark pairs.
In result, all input data of the orthogonal projection --- i.e., the $k$ nearest neighbors from $L$ together with the distances, scores, and 2D versions of the landmarks --- can be cached in shared memory.

The intermediate sub-results represented by $2\times3$ matrices are successively added into privatized copies to avoid explicit synchronization.
The private copies are aggregated at the end using a standard parallel reduction, enhanced with warp-shuffle instructions (a very similar privatization and reduction scheme was used in our previous work~\cite{krulis2020detailed}).

Because the data transfers comprise a considerable portion of the \alg{Shared} algorithm execution time, we have optimized the transfers using alignment and data packing techniques, yielding the \Alg{Aligned} algorithm.
The implementation is based on using vector data types (e.g. \texttt{float4} in CUDA) to enable utilization of $128$-bit load/store instructions, which improves overall data throughput.
The vectorization comes only at a relatively small cost of aligning and padding the vectors to $16$-byte blocks.

\begin{figure}
  \centering
	\includegraphics{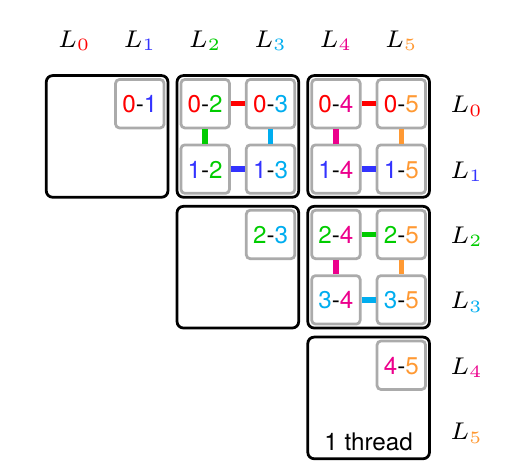}
	\caption{
  Detail of the caching of landmark data in \alg{Registers} projection kernel.
  Multiple landmark pairs (small boxes) are processed by each thread (large boxes).
  Caching of the landmark data in registers allows reuse of loaded data (color lines), thus reducing the amount of memory accesses.
  }
	\label{fig:proj}
\end{figure}

To further improve the data caching, we implemented algorithm \Alg{Registers}, where each thread computes more than one landmark pair in a single iteration, so that the coordinates loaded into registers can be shared as inputs among multiple landmark pair computations.
The data sharing scheme is detailed in Figure~\ref{fig:proj}.
We found that it is optimal to group the threads into small blocks of $2\times2$ computation items, saving half of the data loads.
Larger groups are theoretically possible, but even $3\times3$ caused excessive registry pressure and impaired performance on contemporary GPUs.
The innermost loop of the algorithm iterates over $d$, so that only a single \texttt{float4} value per each landmark is kept in registers.

\subsubsection{Parameters and performance of projection step}

\begin{figure}
	\centering
	\includegraphics[width=21pc]{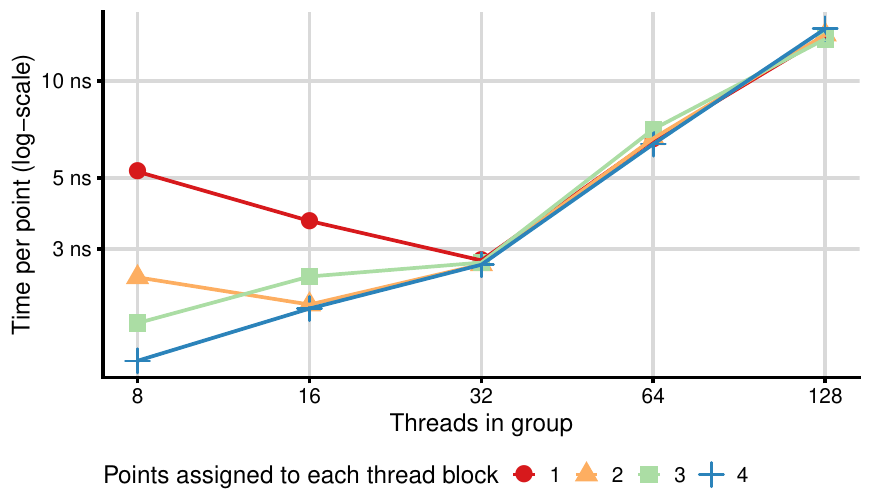}
	\caption{Comparison of various sizes and numbers of thread groups in the \alg{Shared} projection algorithm in the extreme parametrization ($k = 8, d = 4$).}
	\label{fig:multi_block}
\end{figure}

The projection algorithms described in the previous section have only two execution parameters:
The size of the CUDA thread block (i.e., number of threads working cooperatively with shared memory) and the number of data points assigned to a thread block (threads are divided among the points evenly).
We observed that selecting more than one point per thread block is beneficial only in case of relatively small problem instances (low $k$ and $d$), because it prevents under-utilization of the cores.
The effect is illustrated in Figure~\ref{fig:multi_block}.

The optimal size of the CUDA thread blocks depends mainly on the parameters $k$ and $d$.
In case of \alg{Shared} algorithm, optimal values ranged from $8$ (for $k=8$, $d=4$) to $64$ ($k=d=64$).
With the caching optimizations in \alg{Aligned} and \alg{Registers}, the optimal thread block size was slightly higher, reaching $128$ for the most complex problem instances.
We assume this is a direct consequence of the improved memory access efficiency which gives space for parallel execution of additional arithmetic operations.

\begin{figure}
	\centering
	\includegraphics[width=21pc]{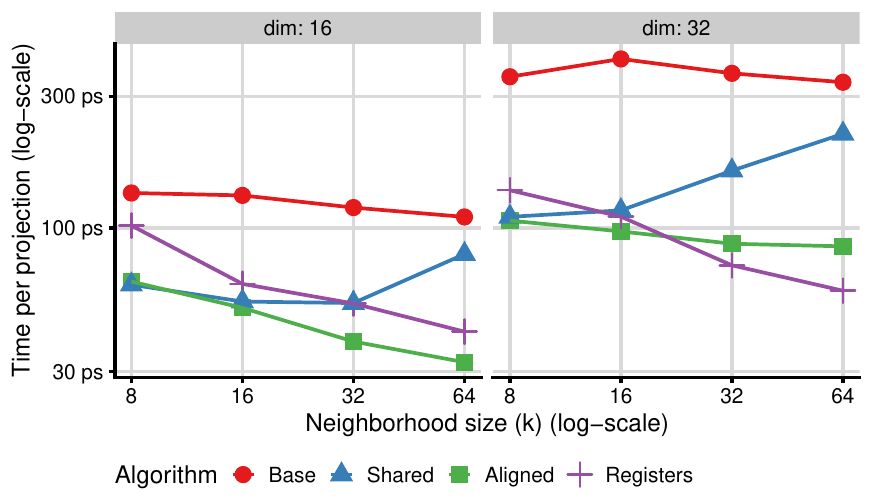}
	\caption{Amortized performance of a single projection operation in the algorithms that compute the projection step, showing the most important problem parametrizations.}
	\label{fig:proj_repre}
\end{figure}

Figure~\ref{fig:proj_repre} shows the performance of the best algorithm configurations for the representative parametrizations.
All three algorithms perform almost equally for small $k$, giving around $3\times$ speedup over \alg{Base}.
The importance of optimizations in \alg{Aligned} and \alg{Registers} grows steadily with increasing of the $k$ parameter, up to around $10\times$ speedup at $k=64$.
In conclusion, the optimal algorithm for the EmbedSOM projection is determined by the dimensionality of the dataset --- \alg{Registers} performs better at higher dimensions ($d\geq32$) while \alg{Aligned} was slightly better for lower dimensions.
Nevertheless, the actual performance difference is negligible and both algorithms can be successfully used as almost optimal.

\subsection{Complete algorithm}\label{sec:impl-complete}

A complete GPU implementation of EmbedSOM algorithm is obtained by combining the best obtained implementations of $k$-NN and projection steps.
In BlosSOM, the selected algorithms \alg{Bitonic} and \alg{Aligned} are simply executed sequentially on large blocks of $X$, sharing only a single data exchange buffer for transferring the $k$-NN data.
Notably, since the data exchange between the algorithm parts is minimal, comprising only of distances and neighbor indexes from the $k$-NN selection, we can assume that no specific optimizations of the interface are required.

Because of the relative complexity of the methods, we did not attempt to compute a theoretically possible data processing throughput.
On the other hand, the collected results seem to scale proportionately to the asymptotic time complexities of the algorithms, roughly following $\mathcal{O}(n\cdot d\cdot g\cdot \log_2 k)$ for the $k$-NN and $\mathcal{O}(n\cdot d \cdot k^2)$ for the projection.
That gives an optimistic outlook on the future scalability of the implementation, especially because the larger problem instances bear no additional overhead.

\begin{figure}
	\centering
	\includegraphics[width=21pc]{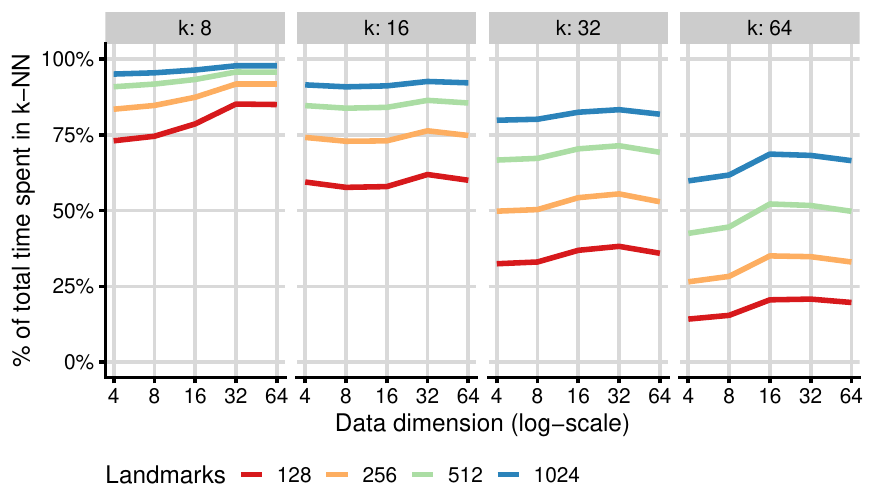}
	\caption{The relative time spent by the $k$-NN computation usually dominates the execution of GPU EmbedSOM, composed of \alg{Bitonic}+\alg{Aligned} algorithms.
  Projection computation time becomes dominant only for relatively impractical parametrizations of low $g$ and high $k$.}
	\label{fig:proj_percent}
\end{figure}

Finally, we highlight the relative computation complexity of both steps, which changes dynamically with $k$, and might be viable as a guide for further optimization.
The results are shown in Figure~\ref{fig:proj_percent}.
We observed that for common parametrizations ($k\simeq 20$, $g\simeq 500$), most of the computation time is spent in $k$-NN step, and projection performance becomes problematic only in cases of almost impractically high $k$.
The performance might be further improved by spatial indexing methods or approximate neighborhood selection algorithms, but we are currently not aware of a scheme that could provide a decisive performance improvement over the optimized brute-force neighbor processing~\cite{krulis2020detailed}.

\section{Results \& Discussion}
\label{sec:results}

\begin{figure}
  \centering
  \includegraphics[width=21pc]{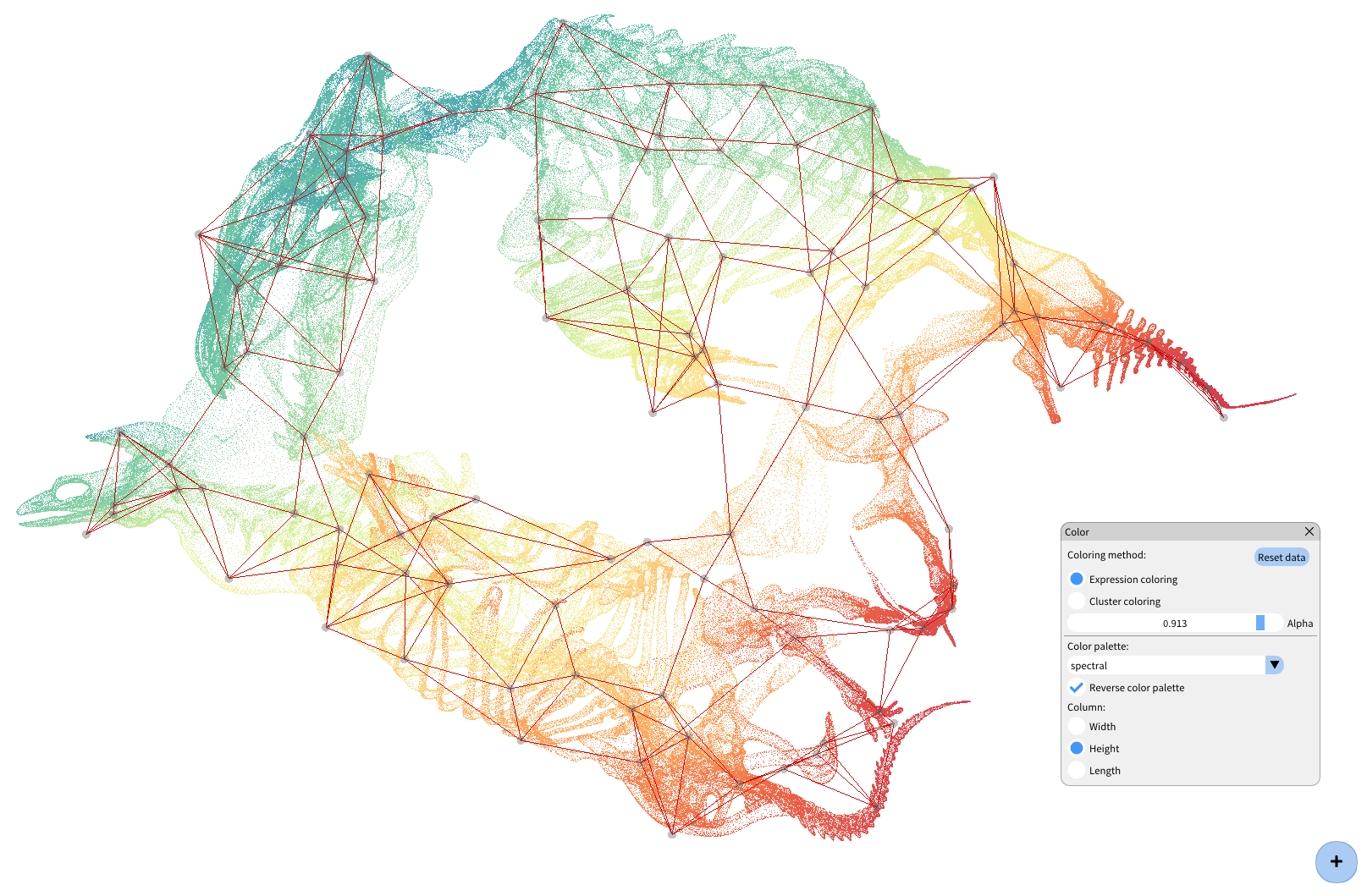}
  \caption{
  Screenshot of the BlosSOM interface during an analysis of the `Tyrannosaurus rex' dataset shows a simplified k-neighborhood graph of high-dimensional landmarks embedded into 2D, the corresponding EmbedSOM projection of the 0.5 million dataset points, and a part of user interface for configuring the colors of the rendered data.
  }
  \label{fig:screenshot}
\end{figure}

We have wrapped the complete implementation of CUDA EmbedSOM in an interactive data analysis environment that implements the semi-supervised methods of dimensionality reduction described in Section~\ref{ssec:dynamic}.
The resulting software, called BlosSOM, is easily capable to utilize the CUDA-compatible hardware to run the workflows on datasets of millions of individual data points.
Performance of the result provides instant, smooth feedback while exploring complex high-dimensional datasets, thus giving and unprecedented tool for comprehending the structure of many datasets.
Figure~\ref{fig:screenshot} shows a screenshot of the main interface, giving the user a view of the `current' projection of data to 2D, together with modifiable positions of 2D landmarks and `settings' windows where the user may customize the data rendering (pictured) and model training parameters.

In this section, we give a summarized overview of the performance, show the applicability of the semi-supervised methods, and discuss the relevance of the analysis on realistic datasets.

\subsection{Accelerated EmbedSOM performance}
\label{ssec:perf}

\begin{figure}
	\centering
	\includegraphics[width=21pc]{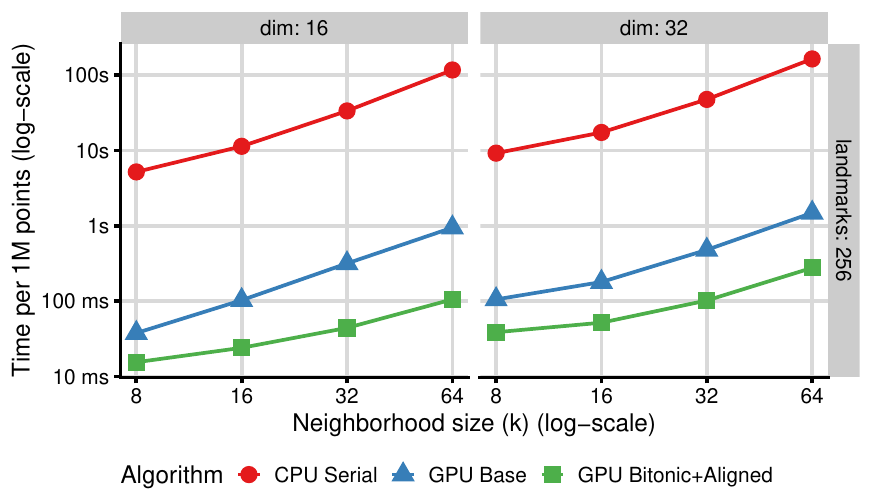}
	\caption{Overview of the achieved dimensionality reduction performance, compared to baseline GPU algorithm and a serial CPU implementation.}
	\label{fig:overall-perf}
\end{figure}

The performance of the optimized implementation of EmbedSOM is one of two main results of this paper.
The summarized measurements are shown in Figure~\ref{fig:overall-perf}:
Our implementation can provide a speedup between $200\times$ and $1000\times$ over a serial CPU implementation, and between $3\times$ and $10\times$ over a straightforward GPU implementation that we used as a baseline.

The performance measurements directly translate to the amount of points per second that may be rendered in an application that utilizes EmbedSOM for dynamic interactive dimensionality reduction.
Considering a target frame rate of 30fps and the usual setting of $k=16$, $g=256$, the method is able to provide real-time rendering of a dataset of 1.1 million 16-dimensional points, or around 0.6 million 32-dimensional points.

Because instant updates are rarely needed at this rate, we may apply a scheme where dataset projection points are updated with smaller frame rate, and displayed with smooth approximations to avoid flicker.
For example, an `update rate' of 3Hz would be sufficient for most scientific applications, allowing full display of around 11 million 16-dimensional points, or 6 million 32-dimensional points.
The data display capacity may be further improved using similar tradeoffs (e.g., dynamic sub-sampling of the data) up until the throughput of the graphical rendering pipeline becomes a bottleneck~\cite{rees2019feature}.

\subsection{Functionality of the supervision methods}
\label{ssec:appl}

\begin{figure}
  \centering
  {\linewidth=21pc
  \begin{tikzpicture}
  \node[inner sep=0] (a) {\includegraphics[width=.66\linewidth]{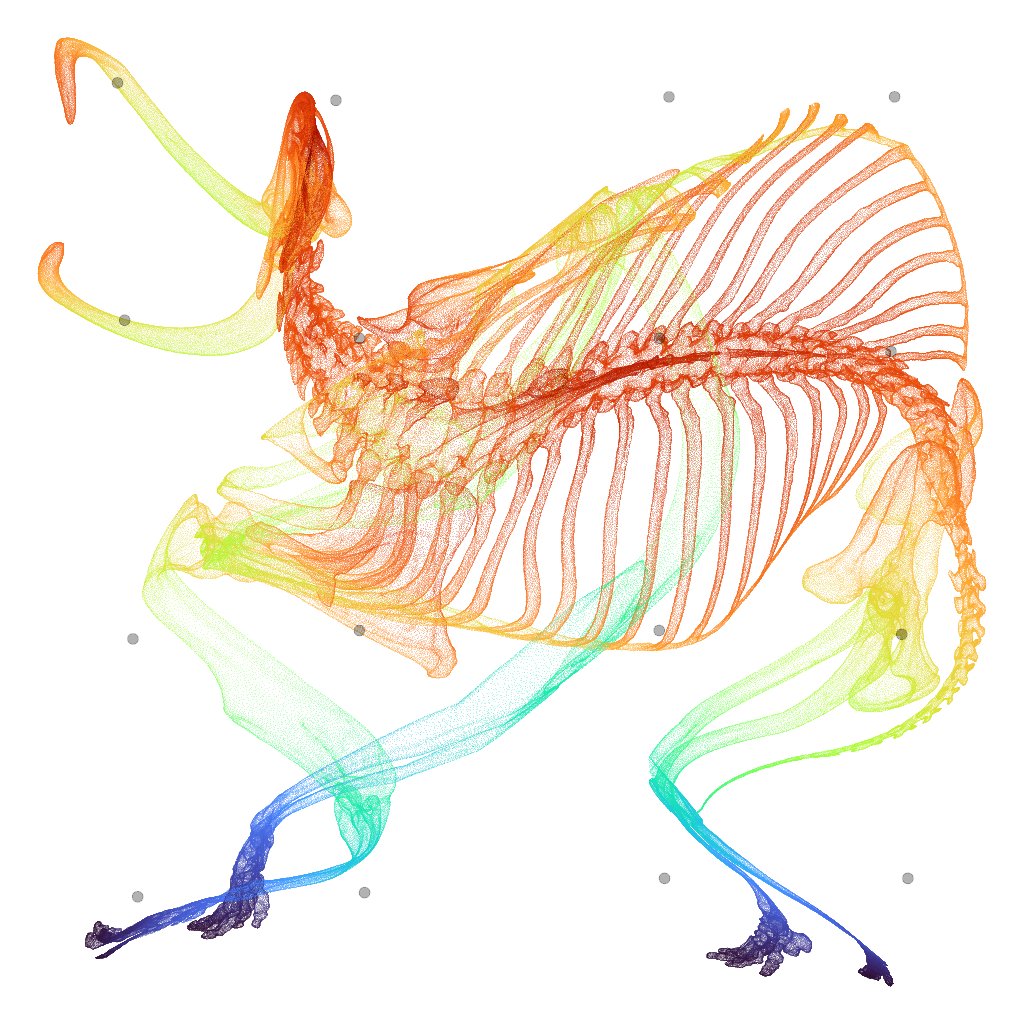}};
  \node[inner sep=0, below=1ex of a] (b) {\includegraphics[width=.66\linewidth]{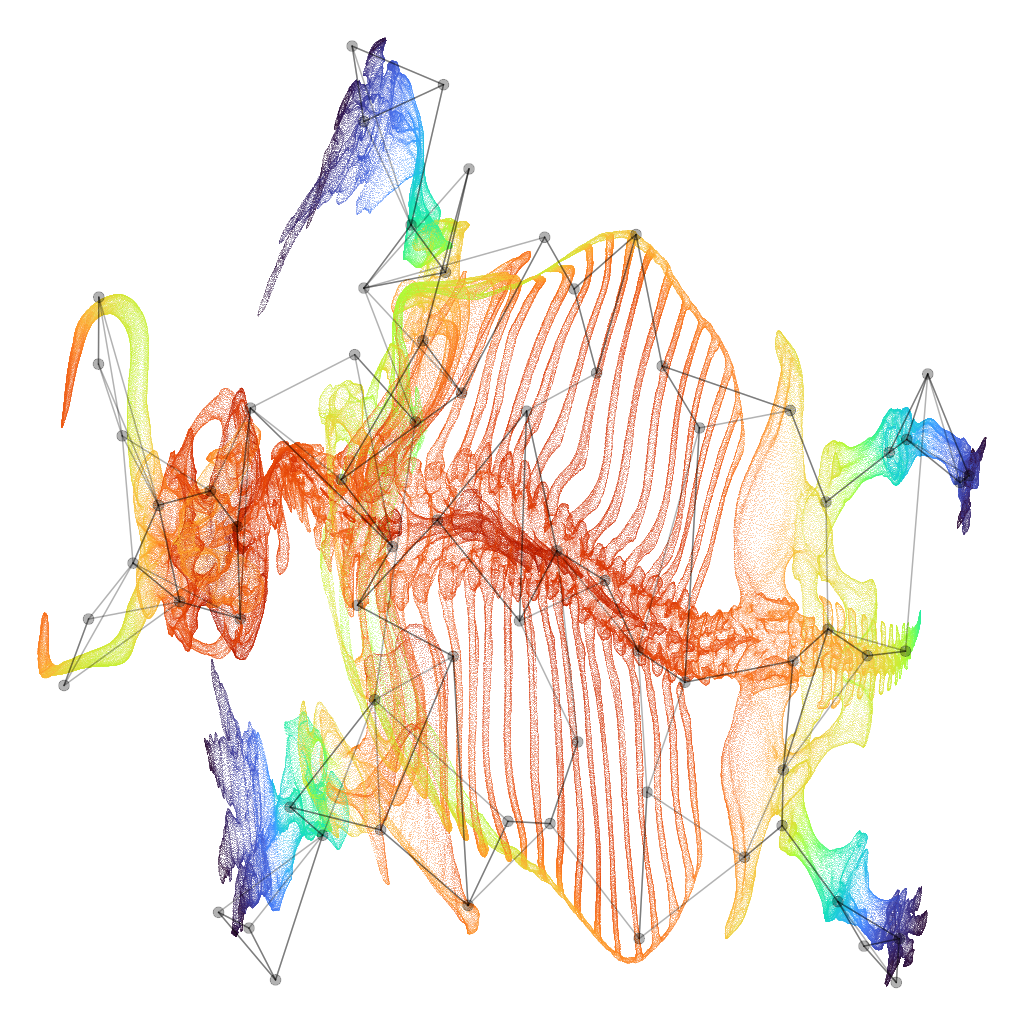}};
  \node[anchor=north east,font=\sf\bfseries] at (a.north west) {A.};
  \node[anchor=north east,font=\sf\bfseries] at (b.north west) {B.};
  \end{tikzpicture}}
  \caption{
  Demonstration of the supervision modes in BlosSOM applied to a 3-dimensional `Mammoth' dataset.
  Data points are colored by physical height to increase clarity.
  Landmarks and their neighborhood graph edges are rendered in transparent gray.
  {\bfseries A.}~Initial display of the data using a small SOM with relatively high $\sigma$.
  {\bfseries B.}~Fine-tuning of the dataset to a specific layout, using additional landmarks organized by the graph approach.
  }
  \label{fig:res-mamut}
\end{figure}

\begin{figure*}
  \centering
  \begin{tikzpicture}
  \node[inner sep=0] (a) {\includegraphics[width=.30\linewidth]{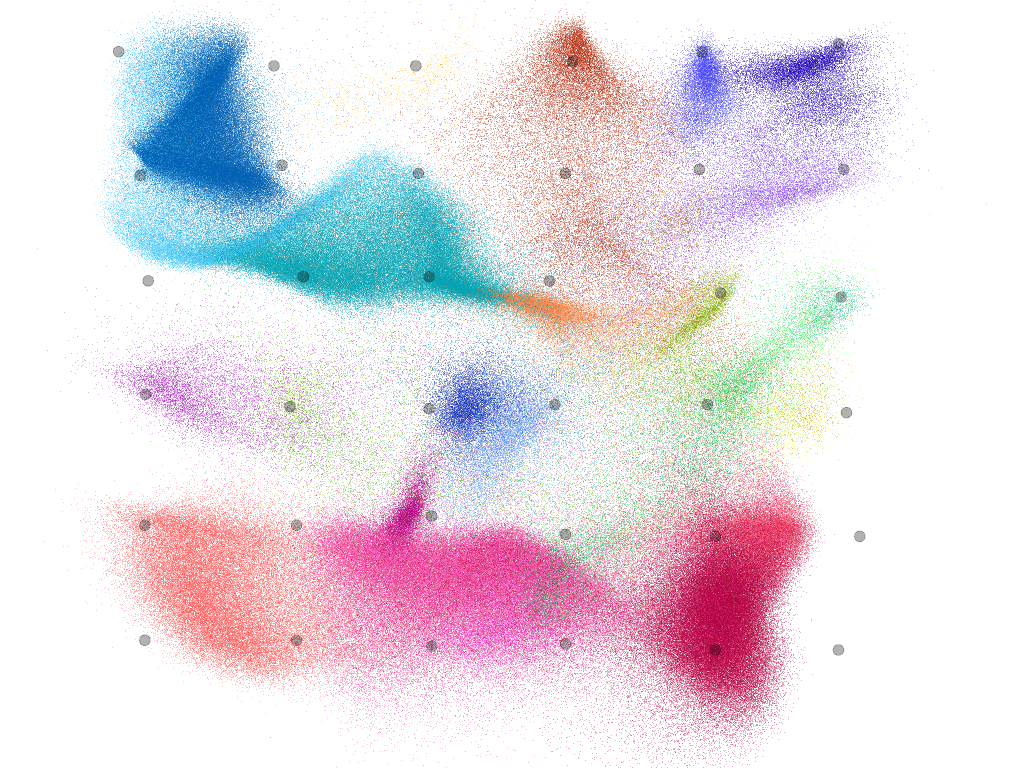}};
  \node[inner sep=0,right=1em of a] (b) {\includegraphics[width=.30\linewidth]{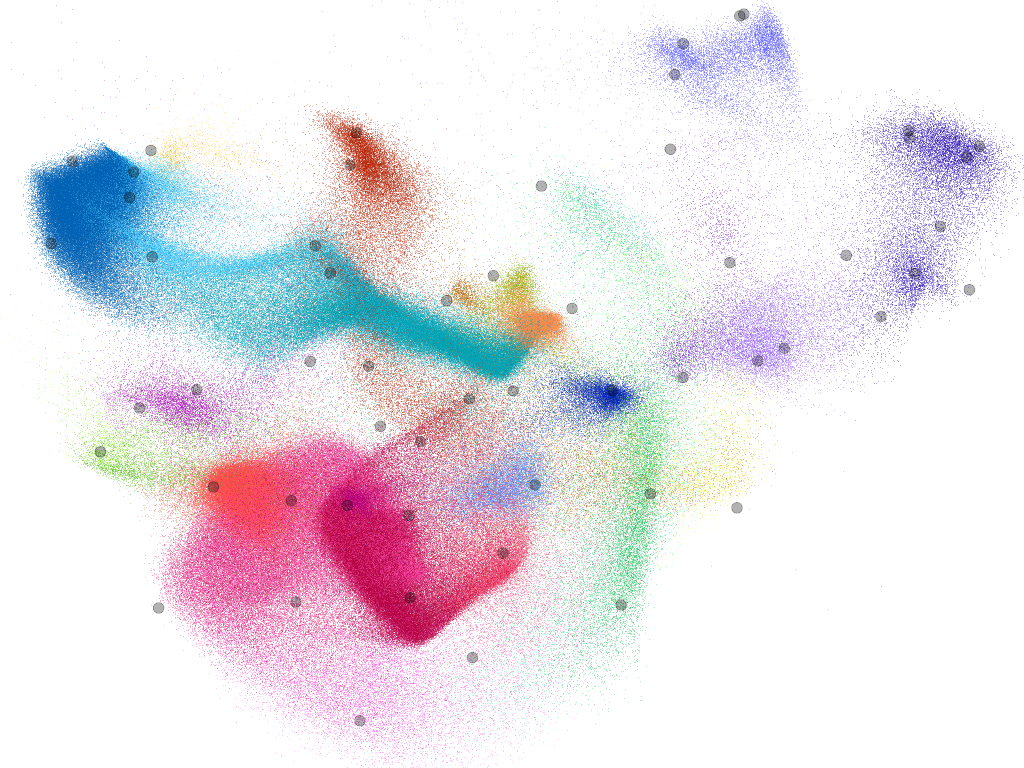}};
  \node[inner sep=0,right=1em of b] (c) {\includegraphics[width=.30\linewidth]{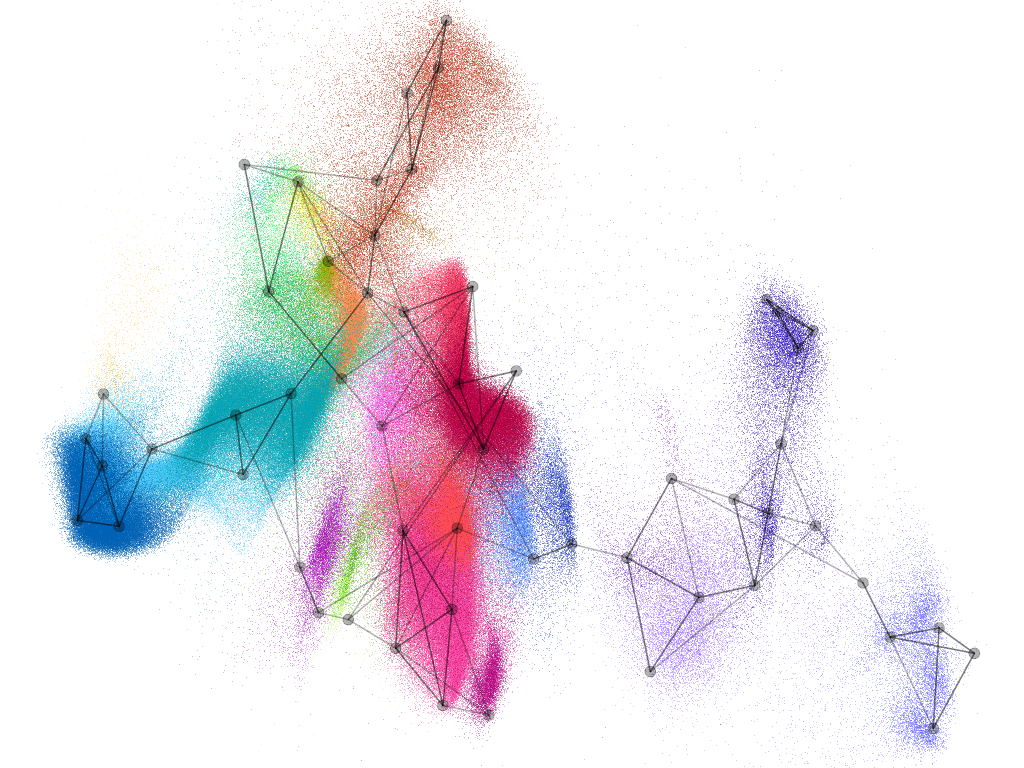}};
  \node[anchor=north west,font=\sf\bfseries] at (a.north west) {A.};
  \node[anchor=north west,font=\sf\bfseries] at (b.north west) {B.};
  \node[anchor=north west,font=\sf\bfseries] at (c.north west) {C.};
  \end{tikzpicture}

  \vspace{2ex}
  \begin{tikzpicture}
  \node[rectangle,draw,inner sep=1pt] (d1) {\includegraphics[width=.1\linewidth]{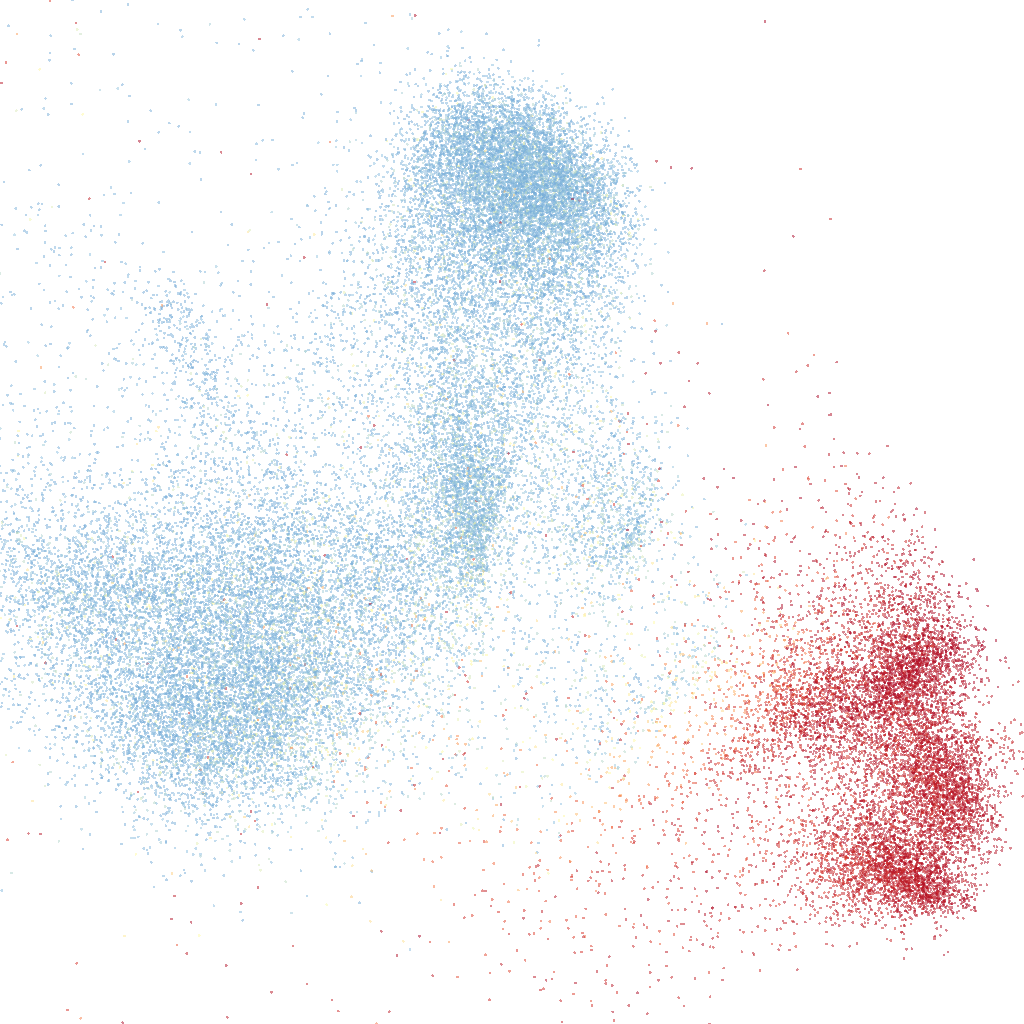}};
  \node[rectangle,draw,inner sep=1pt,right=1em of d1] (d2) {\includegraphics[width=.1\linewidth]{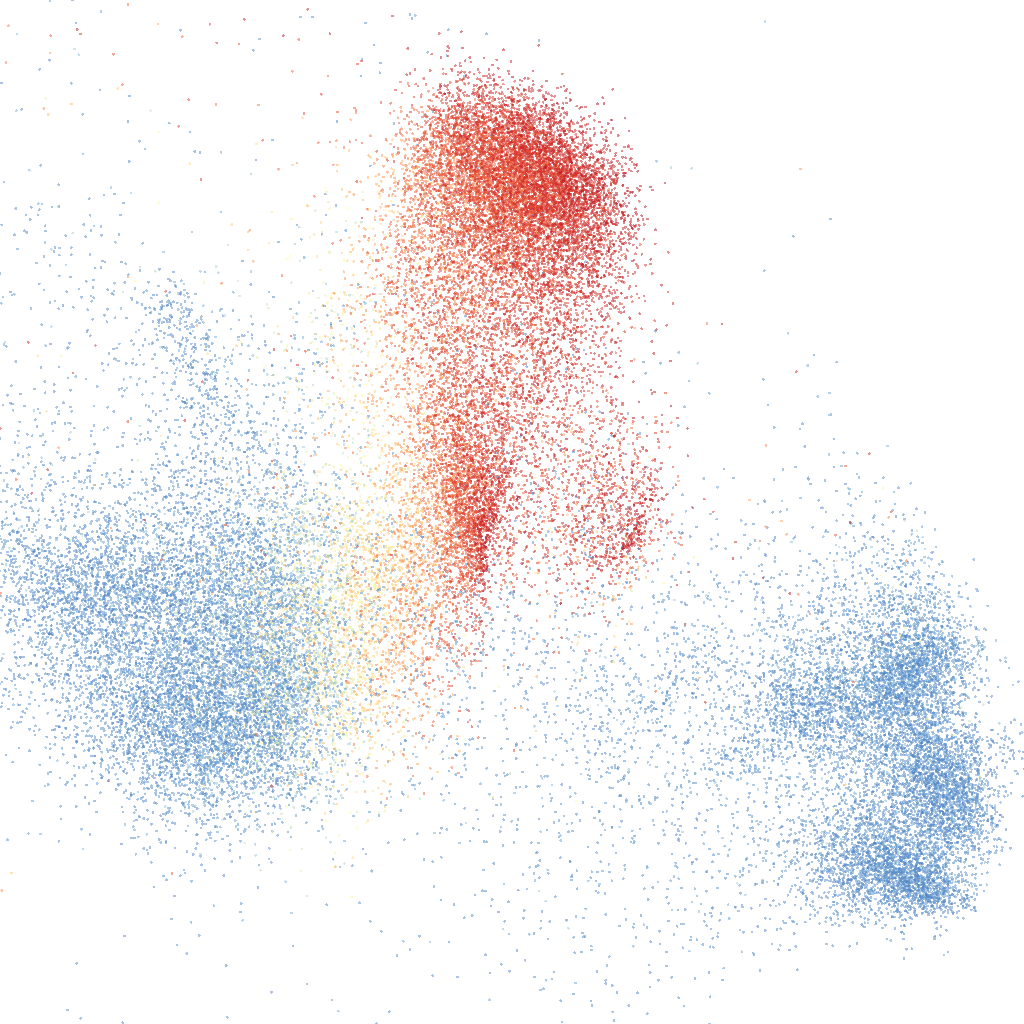}};
  \node[rectangle,draw,inner sep=1pt,right=1em of d2] (d3) {\includegraphics[width=.1\linewidth]{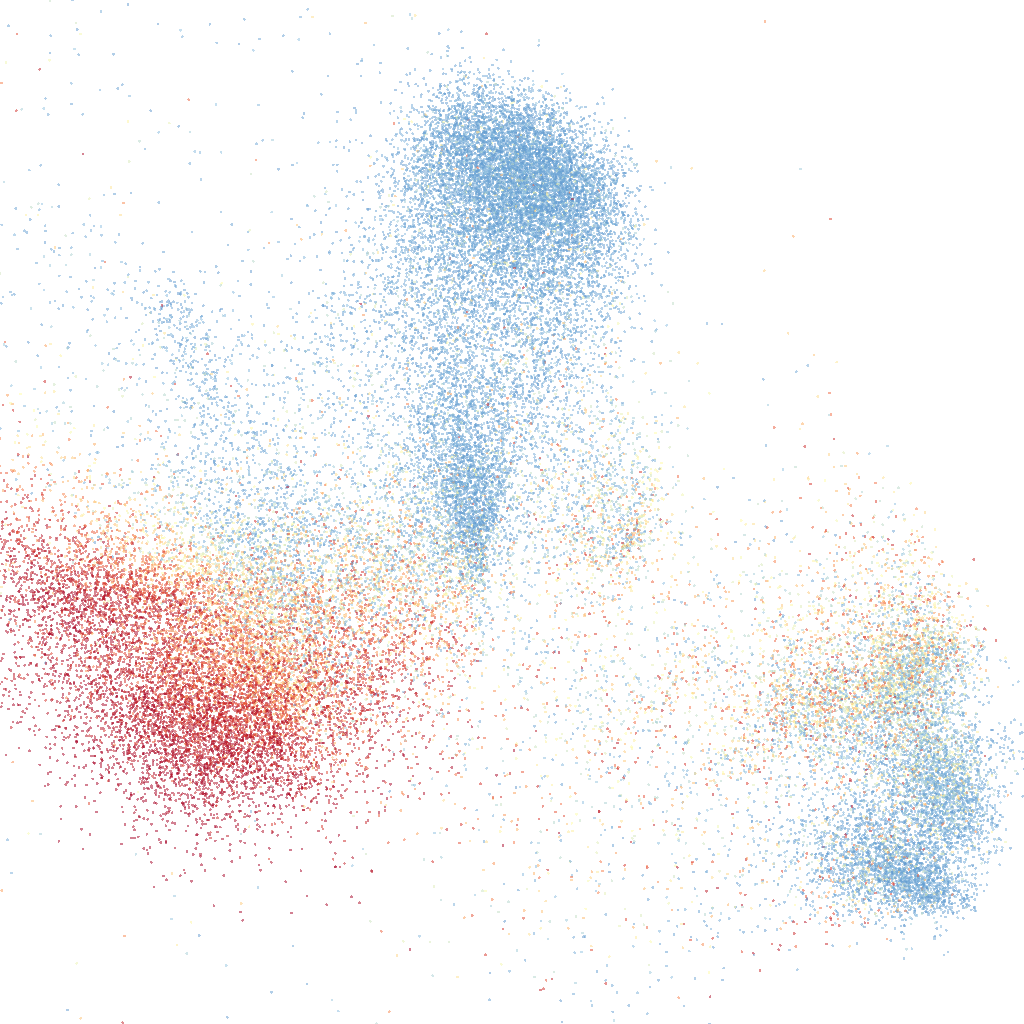}};
  \node[rectangle,draw,inner sep=1pt,right=1em of d3] (d4) {\includegraphics[width=.1\linewidth]{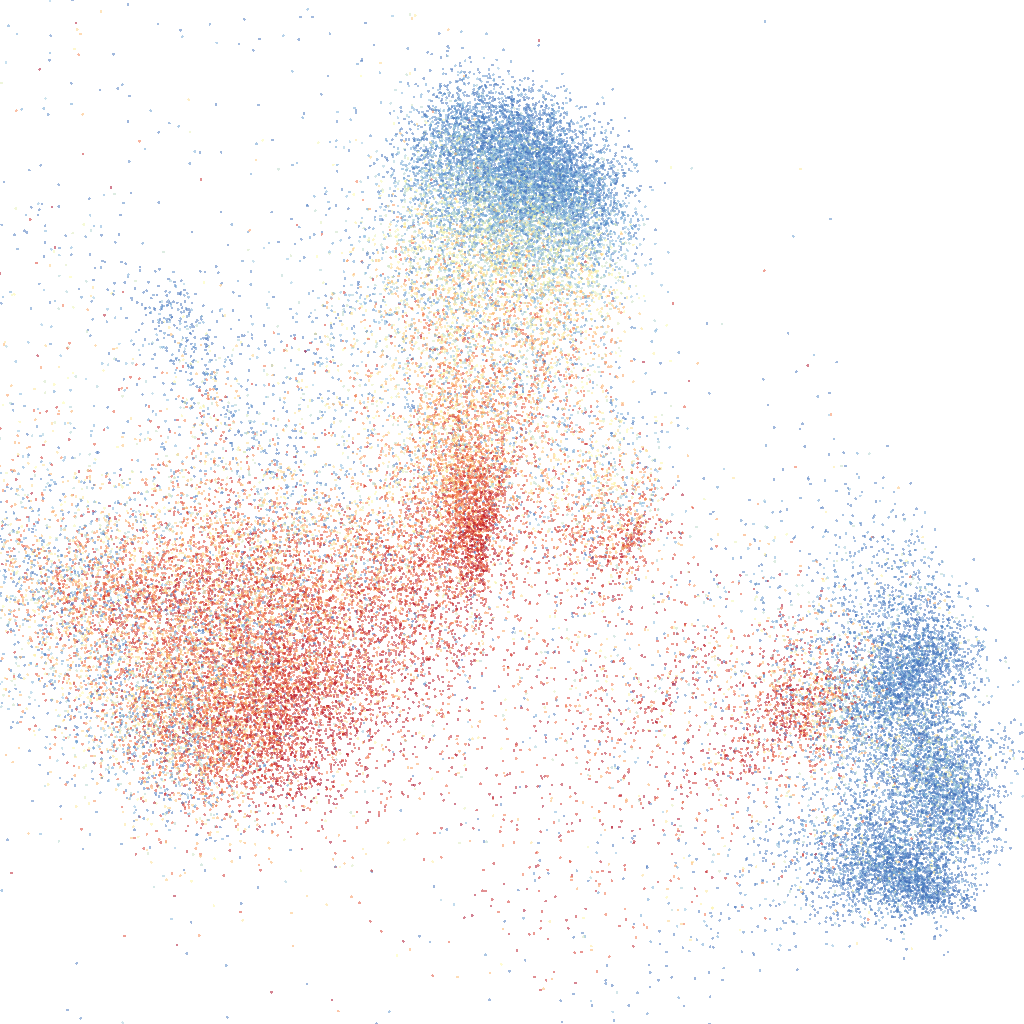}};
  \node[rectangle,draw,inner sep=1pt,right=1em of d4] (d5) {\includegraphics[width=.1\linewidth]{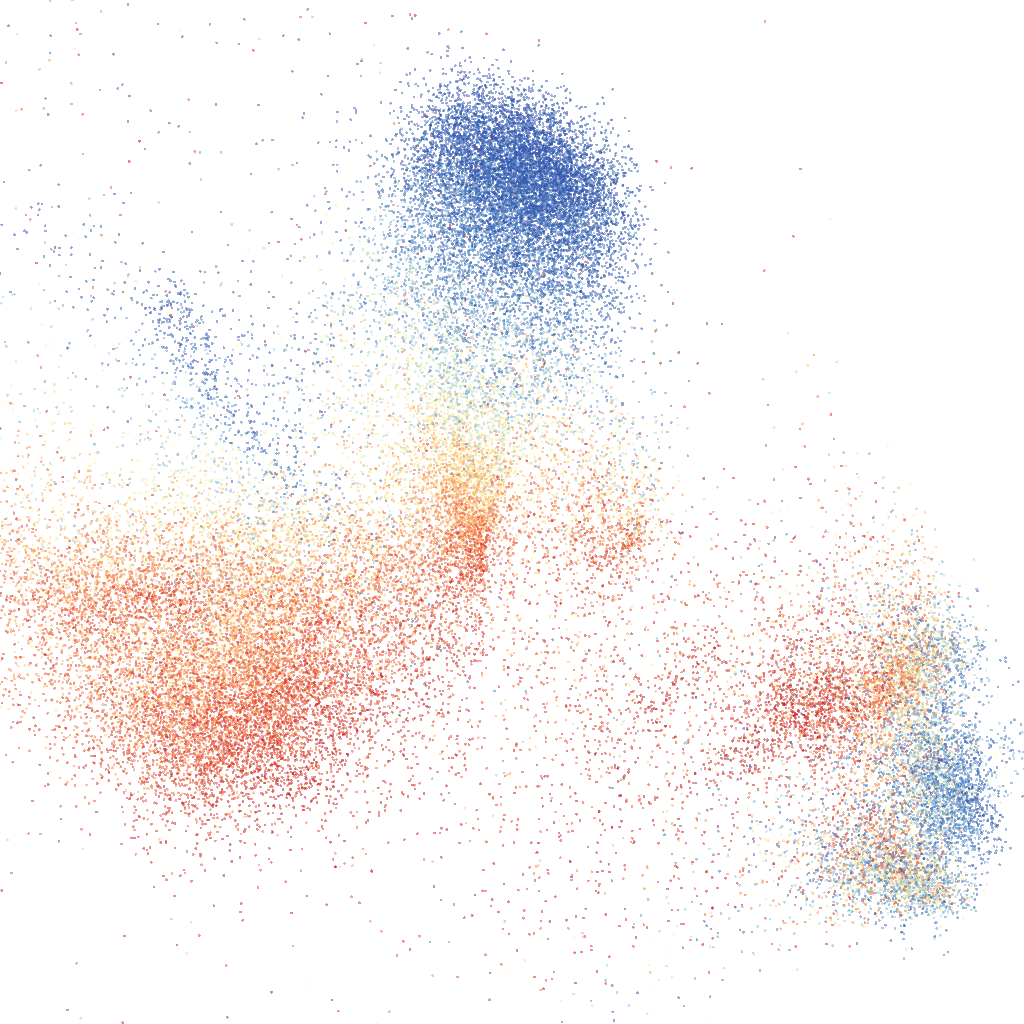}};
  \node[rectangle,draw,inner sep=1pt,right=1em of d5] (d6) {\includegraphics[width=.1\linewidth]{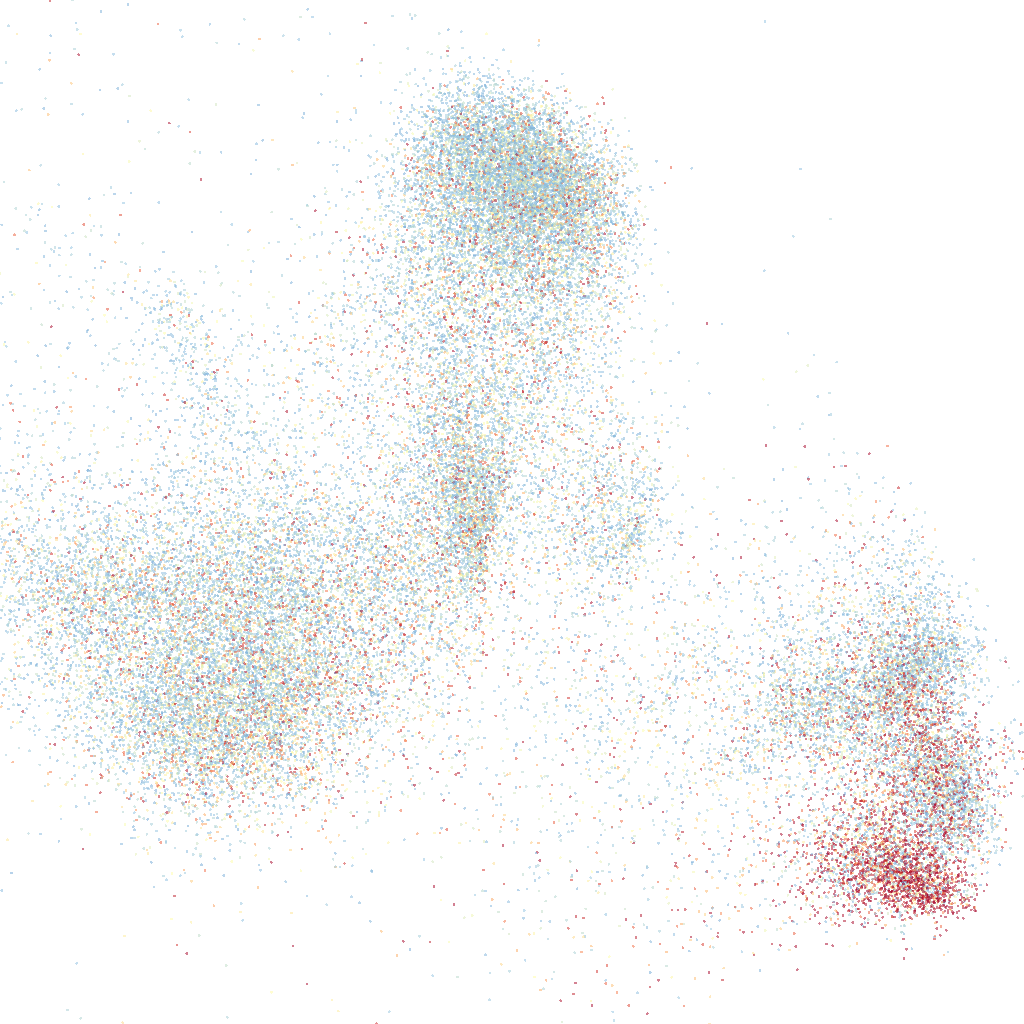}};
  \node[rectangle,draw,inner sep=1pt,right=1em of d6] (d7) {\includegraphics[width=.1\linewidth]{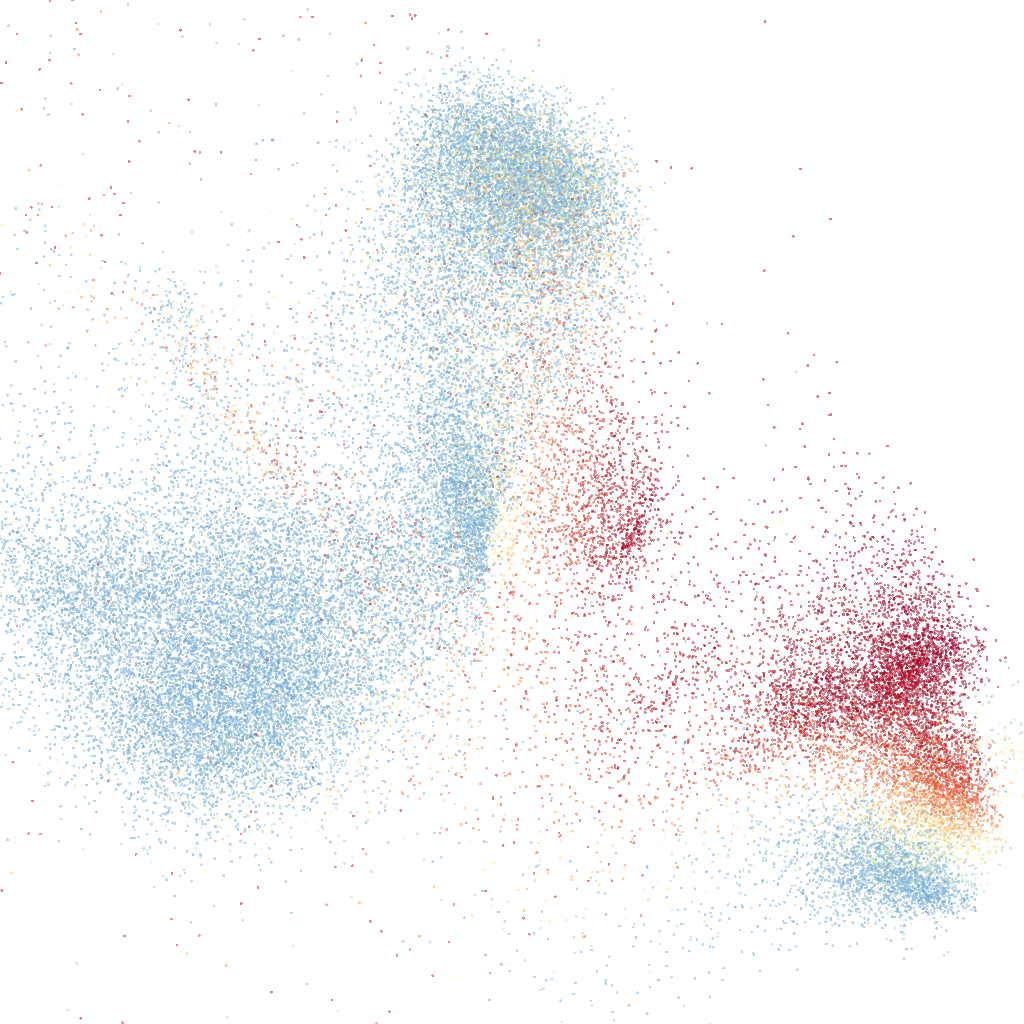}};
  \node[anchor=north east,font=\sf\bfseries] at (d1.north west) {D.};
  \node[below=0ex of d1, font=\sf\scriptsize] {CD4};
  \node[below=0ex of d2, font=\sf\scriptsize] {CD8};
  \node[below=0ex of d3, font=\sf\scriptsize] {CD49b};
  \node[below=0ex of d4, font=\sf\scriptsize] {Ly6C};
  \node[below=0ex of d5, font=\sf\scriptsize] {CD44};
  \node[below=0ex of d6, font=\sf\scriptsize] {CD25};
  \node[below=0ex of d7, font=\sf\scriptsize] {Sca1};
  \end{tikzpicture}
  \caption{
  Demonstration of the exploration and semi-supervised dimensionality reduction of the 39-dimensional `Samusik' dataset using BlosSOM.
  {\bfseries A.}~Initial display of the data using a small SOM; data points are colored as in Figure~\ref{fig:samusik}, landmarks are in gray. A population of interest (T cells) is in the upper right corner, colored in shades of blue-violet.
  {\bfseries B.}~View of the dataset after adding landmarks and manually separating the T cells. Landmark layout is partially optimized by the t-SNE algorithm.
  {\bfseries C.}~Improved separation and positioning of the cell clusters generated using the graph layouting.
  {\bfseries D.}~Detailed view of specific marker expression in the T cells shows subpopulations of biological interest.
  }
  \label{fig:res-samusik}
\end{figure*}

We applied the semi-supervised dimensionality reduction methods implemented in BlosSOM (detailed in Section~\ref{ssec:dynamic}) to several realistic datasets, in order to assess and demonstrate the relevance for data visualization.

Importantly, both described modes of supervision produced relevant results on all testing datasets.
We demonstrate the basic properties of both methods in Figure~\ref{fig:res-mamut}, using a 3-dimensional dataset of a mammoth skeleton as an easily comprehensible example.
We observed that the supervised self-organization approach (Section~\ref{sssec:sup-som}) quickly provides a good view of complicated datasets, even on self organizing maps that are comparatively small.
In case of the mammoth dataset, even a 4\texttimes4 self-organizing map was able to provide an interesting reorganization of the dataset (Figure~\ref{fig:res-mamut}A), giving a reasonable view to both the mammoth chest and the exteriors (legs and tusks).
In all experiments, the topology reconstruction approach (Section~\ref{sssec:sup-graph}) was more useful to fine-tune the layout learned by the self-organizing map, allowing us to reorganize parts of the dataset and add detail with additional landmarks.
For demonstration, we utilized it to rework the layout of the mammoth visualization to neglect the vertical structure (Figure~\ref{fig:res-mamut}B), which is complicated with the self-organizing approach.

The experiments showed an important benefit of the supervision:
In many cases, the unsupervised methods make correct but undesirable assumptions about dataset layout, and the generic `portable' choices in the algorithms actually limit the applicability to specific cases.
For example, in the mammoth layout in Figure~\ref{fig:res-mamut} the self-organizing map attempts to produce the best fit of the animal to a square topology, resulting in unnatural positioning and deformation of limbs.
In the semi-supervised setting, user is free to modify the self-organizing map to a rough shape of the expected data, improving both the quality of the SOM fit and the comprehensibility of the visualization.
The possibility to customize the layout of data is further beneficial with datasets where the correct projection to 2D is ambiguous or impossible.
For example, supervision allows the user to choose a representation where the ambiguity is solved by adherence to a commonly accepted data layout guidelines, and the impact of non-planarity is minimized by moving the data overlaps away from regions of interest.

\subsection{Application to single-cell cytometry data}

To show the utility for biological use-cases, we used BlosSOM to visualize a dataset by Samusik~et~al.~\cite{samusik2016automated} that was used to map the structure and development of immune cell populations within the bone marrow.
The dataset is interesting because it contains 24 unique cell populations originally identified by domain experts, and most of these populations possess further subpopulations, and are often connected by cell development pathways.

Figures~\ref{fig:res-samusik}A--\ref{fig:res-samusik}C show the process of exploring visualizing the dataset in BlosSOM.
For clarity, we colored the visualizations by unique colors per each identified population.
As in previous case, we started the exploration with generating a rough SOM-based projection of data, followed by manual modifications to landmark positions and optimization of their layout using t-SNE and graph embedding.
In this particular case, we picked the population of the T cells (that can be easily identified in BlosSOM by the presence of CD3 marker) that we separated from the other clusters in order to detail its contents, for which the graph-based layout provided best opportunity.

The result enables the users to identify more separate and biologically relevant cell sub-populations, as seen on Figure~\ref{fig:res-samusik}D.
In this case, the CD4\textsuperscript{+} T cells are clearly subdivided into the populations of activated and memory cells based on the expression of CD44, CD25, and Sca1 markers.
The CD8\textsuperscript{+} T cell population is visibly split into naive and memory cells, based on the expression of CD44 and Ly6C markers.
Similar observations have been reported by the authors of the dataset~\cite{samusik2016automated}.

Notably, the demonstrated exploration and visualization process partially relieves the problematic choice of detail level of data visualization in single cell cytometry (the `global and local' properties of datasets):
Users are typically forced to decide between visualizations that optimize either the display of more coarse-grained dataset properties (such as positioning of cluster groups) or fine-grained properties (such as shape of sub-clusters and positioning of small populations), but not both.
We believe that the approach above poses a valid alternative solution for this problem.
In particular, the ability of the user to pick features that should be detailed (and neglect ones with no importance for a given narrative) may improve the ability of the scientists to produce highly focused visualizations for specific areas of science, possibly increasing practical and educative value of their output.

\subsection{Future work}
\label{ssec:future}
While the current version of BlosSOM is already suitable for exploration of many complicated high-dimensional datasets, we believe that materializing the discovery by scientist users might increase the utility in dataset annotation and dissection.
In particular, we expect that e.g.~a `cluster brushing' tool (known from many other data analysis tools) could give the scientist a highly efficient and intuitive way to annotate the discovered clusters, which then could serve as a reference dataset for unbiased approaches based on supervised clustering and machine learning.

While the currently implemented methods of supervision are sufficient for many dataset types, many additional methods are possible.
For example, since we already implemented t-SNE as a landmark layouting algorithm, we expect that many other dimensionality reduction algorithms, including UMAP and TRIMAP, will be useful for the same purpose, given a `smooth' implementation can be found.
Similarly, other graph generation and layout methods may be more useful for the topology reconstruction approach, including e.g. spanning trees and U-matrix-style SOM neighborhood graphs.

\section{Conclusion}
\label{sec:outro}

We have presented BlosSOM, a novel software for semi-supervised dimensionality reduction and visualization of large datasets.
BlosSOM utilizes a GPU-accelerated implementation of the EmbedSOM algorithm as a highly efficient base for projecting the data to 2D, and improves its use with several supervision methods that allow the users to interactively and intuitively steer the process towards the desired solution with feedback.

The GPU implementation in CUDA was thoroughly benchmarked and optimized.
In BlosSOM, it is used to dynamically project the data points in an interactive visualization environment, where it re-projects and re-renders all data points every frame at high frame rate.
On typical datasets, the optimized version is able to project more than 1 million of individual data points each frame, while maintaining a frame rate of 30fps or higher.

We described and implemented several methods for user interaction with the landmark-based data model in EmbedSOM, based on self-organizing maps and graph embedding.
The combination of the approaches in BlosSOM provided a solution to several challenges typically encountered with unsupervised dimensionality reduction.
Finally, we demonstrated the use of BlosSOM on a biologically relevant use-case from single-cell cytometry, where it gives an effective way to produce desirable visualizations with variable level of details.

We believe that the presented methodology will find use in explorative analysis of complex datasets, and provide a base for constructing intuitive, user-friendly annotation and dissection tools for single-cell cytometry data.

\subsection{Data and software availability}\label{ssec:data}

BlosSOM is available as free and open-source software from \texttt{https://github.com/\-molnsona/\-blossom}.
Benchmark results are available from \texttt{https://github.com/\-asmelko/\-embed\-som-bench\-marks}.

Datasets displayed in Sections \ref{sec:methods} and \ref{sec:results} are available from FlowRepository (\texttt{http://flow\-repository.org/\-id/FR-FCM-ZZPH}, file \texttt{Samusik\_all.fcs}) and from Smithsonian institute 3D digitization repository (\texttt{https://3d.si.edu/}, datasets `Mammuthus primigenius (Blumbach)' and `Tyrannosaurus rex', the 3D point coordinates were extracted manually from the vertex coordinates available in \texttt{.obj} files).

\section*{Acknowledgments}
This work was supported by
by ELIXIR~CZ LM2018131~(MEYS),
Czech Science Foundation project \mbox{19-22071Y}~(GAČR),
by Charles University grant \mbox{SVV-260451},
by European Regional Development Fund and the state budget of the Czech Republic, project AIIHHP, Grant/Award Number CZ.02.1.01/0.0/0.0/16\_025/0007428 OP RDE (MEYS),
and by Project for Conceptual Development of Research Organization No.~00023736 from the Ministry of Health of the Czech Republic.

\printbibliography

\end{document}